\documentclass{IEEEtran}
\IEEEoverridecommandlockouts
\usepackage{cite}        

\makeatletter
\def\@cite#1#2{{\color{blue}[{#1\if@tempswa , #2\fi}]}}
\makeatother

\usepackage{booktabs}
\usepackage{amsmath,amssymb,amsfonts,amsthm}
\newtheorem{theorem}{Theorem}
\newtheorem{Lemma}{Lemma}
\theoremstyle{definition} 

\usepackage{algorithm}
\usepackage{algpseudocode}
\usepackage{adjustbox} 
\usepackage{graphicx}
\usepackage{subcaption}
\usepackage{textcomp}
\usepackage{tabu} 
\usepackage[x11names]{xcolor} 
\usepackage[colorlinks=true, linkcolor=blue, citecolor=blue, urlcolor=blue]{hyperref} 
\def\BibTeX{{\rm B\kern-.05em{\sc i\kern-.025em b}\kern-.08em
    T\kern-.1667em\lower.7ex\hbox{E}\kern-.125emX}}
\begin{document}
\allowdisplaybreaks 
\title{Efficient UAV Swarm-Based Multi-Task Federated Learning with Dynamic Task Knowledge Sharing\\
}

\author{\IEEEauthorblockN{Yubo Yang,}
\and
\IEEEauthorblockN{Tao Yang,}
\IEEEauthorblockA{\textit{Member, IEEE,}}
\and
\IEEEauthorblockN{Xiaofeng Wu,}
\and
\IEEEauthorblockN{Ziyu Guo,}
\IEEEauthorblockA{\textit{Member, IEEE,}}
\and
\IEEEauthorblockN{and Bo Hu,}
\IEEEauthorblockA{\textit{Member, IEEE}}
}

\maketitle

\begin{abstract}
Unmanned aerial vehicle (UAV) swarms are widely used in emergency communications, area monitoring, and disaster relief. Coordinated by control centers, they are ideal for federated learning (FL) frameworks. However, current UAV-assisted FL methods primarily focus on single tasks, overlooking the need for multi-task training. In disaster relief scenarios, UAVs perform tasks such as crowd detection, road feasibility analysis, and disaster assessment, which exhibit time-varying demands and potential correlations. In order to meet the time-varying requirements of tasks and complete multiple tasks efficiently under resource constraints, in this paper, we propose a UAV swarm based multi-task FL framework, where ground emergency vehicles (EVs) collaborate with UAVs to accomplish multiple tasks efficiently under constrained energy and bandwidth resources. Through theoretical analysis, we identify key factors affecting task performance and introduce a task attention mechanism to dynamically evaluate task importance, thereby achieving efficient resource allocation. Additionally, we propose a task affinity (TA) metric to capture the dynamic correlation among tasks, thereby promoting task knowledge sharing to accelerate training and improve the generalization ability of the model in different scenarios. To optimize resource allocation, we formulate a two-layer optimization problem to jointly optimize UAV transmission power, computation frequency, bandwidth allocation, and UAV-EV associations. For the inner problem, we derive closed-form solutions for transmission power, computation frequency, and bandwidth allocation and apply a block coordinate descent method for optimization. For the outer problem, a two-stage algorithm is designed to determine optimal UAV-EV associations. Furthermore, theoretical analysis reveals a trade-off between UAV energy consumption and multi-task performance, characterized by an \(\mathcal{O}(\sqrt{V}, 1/V)\) relationship. Simulation results verify the effectiveness of the algorithm.
\end{abstract}


\section{Introduction}
Unmanned Aerial Vehicle (UAV)-assisted networks are seen as an important enabler for post-5G and upcoming 6G networks. Compared with other equipment, UAVs have greater flexibility and a wider field of view, making them suitable for emergency communications and wide-area on-demand data collection\cite{b0}. In particular, they can serve as aerial mobile base stations (BS) in remote or hard-to-reach areas, or act as temporary infrastructure in regions where traditional network infrastructure has been damaged by natural disasters such as earthquakes and tsunamis. Additionally, UAVs can be used to train popular machine learning models onboard, which is critical for tasks such as trajectory planning and object recognition\cite{b1}\cite{b14}. In a multi-UAV system, UAVs operate in a coordinated manner to support reliable and efficient communications or to complete specific tasks\cite{b3}, with this coordination typically being assisted by ground base stations. For example, in emergency rescue scenarios with poor network coverage, a group of UAVs are controlled by a ground emergency vehicle (EV). However, UAVs are usually deployed in different areas and often collect data dynamically, it is unrealistic to continuously transmit raw data from each UAV to the EV or to frequently travel between the accident site and the EV for charging, due to the limited communication resources and energy capacity of UAVs\cite{b4}. Faced with these challenges, federated learning (FL)\cite{b5} has emerged as a promising distributed data processing and model training framework. By leveraging the onboard training capabilities of UAV platforms, it enables multiple UAVs to collaboratively train models without transmitting a large amount of on-board sensed data, thereby alleviating communication costs and improving the efficiency of model training.

Numerous studies have explored UAV-assisted federated learning scenarios, which can be broadly categorized into two types: UAVs serving as aggregation servers \cite{b15} and UAVs performing on-board training \cite{b16}. For the former, UAVs receive parameters uploaded by ground devices and aggregate them for broadcast. For instance, Yang et al.\cite{b17} and Do et al.\cite{b18} studied the UAV deployment and resource allocation to accelerate the model convergence under resource constraints. Tang et al.\cite{b19} proposed a UAV-enabled wireless power communication network. In their framework, UAVs not only aggregate FL model parameters but also charge UEs via wireless power transfer technology. To ensure reliable communication, Wang et al. \cite{b20} proposed a two-layer hierarchical FL scheme, using UAVs as relays between FL parameter servers and local clients. For the latter scenario, Li et al.\cite{b21} introduced a UAV object detection task framework based on online FL and adopted Over-the-Air Compotation (OTA)\cite{b22} technology to improve transmission efficiency. Shen et al.\cite{b23} studied how to minimize the overall training energy consumption of UAV swarms by jointly optimizing local convergence thresholds, computing resources, and bandwidth allocation. In order to solve the information asymmetry and incentive mismatch problems between UAVs and model owners, Lim et al.\cite{b24} designed a UAV swarm FL framework based on multi-dimensional contract theory. Similarly,
He et al.\cite{b25} proposed a cluster FL architecture that clusters UAV swarms based on their optimization directions, while a three-stage Stackelberg game was formulated to optimize the allocation of limited resources. Jiang et al. \cite{b26} combined Digital Twin (DT) with edge computing and established an AirComp-enabled FL architecture in UAV swarms for communication efficient DT modeling and convergence analysis.

All the studies mentioned above focus on single-task learning, where multiple UAVs collaborate to accomplish the same task. However, UAVs are often deployed across different regions to sense the target object or situation, which can be leveraged for a variety of tasks. For example, during firefighting operations, UAVs are employed
for situational awareness tasks, identifying scene information
to determine the safest direction for extinguishing the fire.
They are also used to monitor the fire source area, enabling
effective fire suppression. These tasks may exhibit underlying correlations that could be exploited for improved performance. Several studies have explored multi-task federated learning. For example, Smith et al. \cite{b27} proposed the MOCHA algorithm, utilizing multi-task learning to tackle data heterogeneity in federated learning. It treats each client's model as an independent task, learning personalized models while requiring uniform parameter dimensionality across all clients. Similarly, Ma et al. \cite{b28} introduced the  Over-the-Air Federated Multi-Task Learning framework, in which multiple learning tasks are deployed on edge devices. However, these tasks originate from different datasets and the framework requires uniform parameter dimensionality across tasks as well. 
There are also some other works that treat multi model personalization as multi-task\cite{b29}\cite{b30}.
Additionally, several studies have focused on a concept known as multi-job federated learning. For instance, Yang et al. \cite{b8} introduced a multi-job vertical federated learning framework, in which multiple independent models are simultaneously trained in cross-island environments, yet they have overlooked the potential correlations between the different jobs. Wang et al. \cite{b31} employed a multi-armed bandit framework to jointly optimize the training efficiency of multiple jobs in dynamic and stochastic environments. Shi et al. \cite{b32} investigated the fairness-aware job scheduling problem in multi job federated learning. Research in this field has primarily focused on reducing client idle time and improving the parallelism of task training, rather than leveraging task correlations to enhance task performance.

When correlations exist among multiple tasks, some studies have focused on jointly learning these tasks and leveraging inter-task correlations to improve overall performance. Nevertheless, these approaches are generally restricted to centralized training settings. For example, Long et al. \cite{b34} proposed a Multi-Linear Relationship Network, which jointly learns transferable features and the multi-linear relationships between tasks and features. Similarly, Jia et al. \cite{b35} trained multiple tasks simultaneously and shared a single feature extractor across all tasks, but they can not address the performance degradation caused by conflicting tasks. Tu et al. \cite{b36} introduced a federated learning framework with a dynamic layer-sharing mechanism, designed to learn personalized models for each client. Also, some studies have shown that sharing feature extractors among related tasks can improve model training speed and robustness \cite{b11}, \cite{b33}, \cite{Caruana}, but they fail to consider the dynamic nature of task correlations and energy consumption issues in practical deployments.

Unlike the studies mentioned above, the UAV swarm based multi-task federated learning considered in this paper involves performing multiple tasks simultaneously using the same dataset collected by the UAVs. Considering the inefficiency of independently learning each task using traditional single-task learning frameworks \cite{b10}, as well as the potential intrinsic correlations and time-varying nature of the tasks, this paper primarily focuses on efficiently identifying the dynamic correlations among multiple tasks in an online setting. Moreover, this paper aims to exploit these correlations to achieve cross-task knowledge sharing by sharing feature extractor updates across related tasks, thereby enhancing the overall performance of all tasks under the constraints of UAV energy consumption and computational resources. The contributions of this paper are summarized as follows:

\begin{itemize}
    \item We propose a novel UAV swarm-based multi task federated learning framework, where multi-view data collected by UAVs from varying altitudes enables simultaneous training of multiple tasks. To address dynamic task requirements, we introduce a task attention mechanism to measure the time varying importance of tasks. Additionally, we incorporate a task affinity metric to capture the dynamic intrinsic correlations among tasks, thereby facilitating task knowledge sharing and improving both task performance and system robustness.
    
    \item To improve the task performance under UAV energy consumption and training time constraints, the Lyapunov optimization is used to mitigate the impact of non-causal information. Subsequently, inner and outer layer problems are constructed. For the former, the trade-off relationship between UAV training and transmission energy consumption under single-round time constraints is elucidated, alongside the derivation of optimal closed form expressions for the UAV's computation frequency, transmission power, and bandwidth allocation. For the latter, a low-complexity optimal two-stage UAV-EV association algorithm is proposed.
  
    \item Theoretical analysis of the proposed algorithm is performed, demonstrating that it satisfies the UAV energy budget under the \(\sqrt{V}\) factor and achieves near-optimal multi-task performance under the \(1/V\) factor. This establishes an \(O(\sqrt{V}, 1/V)\) trade-off between UAV energy consumption and multi-task training performance. Extensive experiments demonstrate that the proposed algorithm can accurately identify dynamic task correlations and outperforms several baseline in both task performance and resource utilization efficiency.
\end{itemize}


\section{System Model}
\subsection{Federated Learning System}
\begin{figure}[h]\label{fig1}
    \centering
    \includegraphics[width=0.45\textwidth]{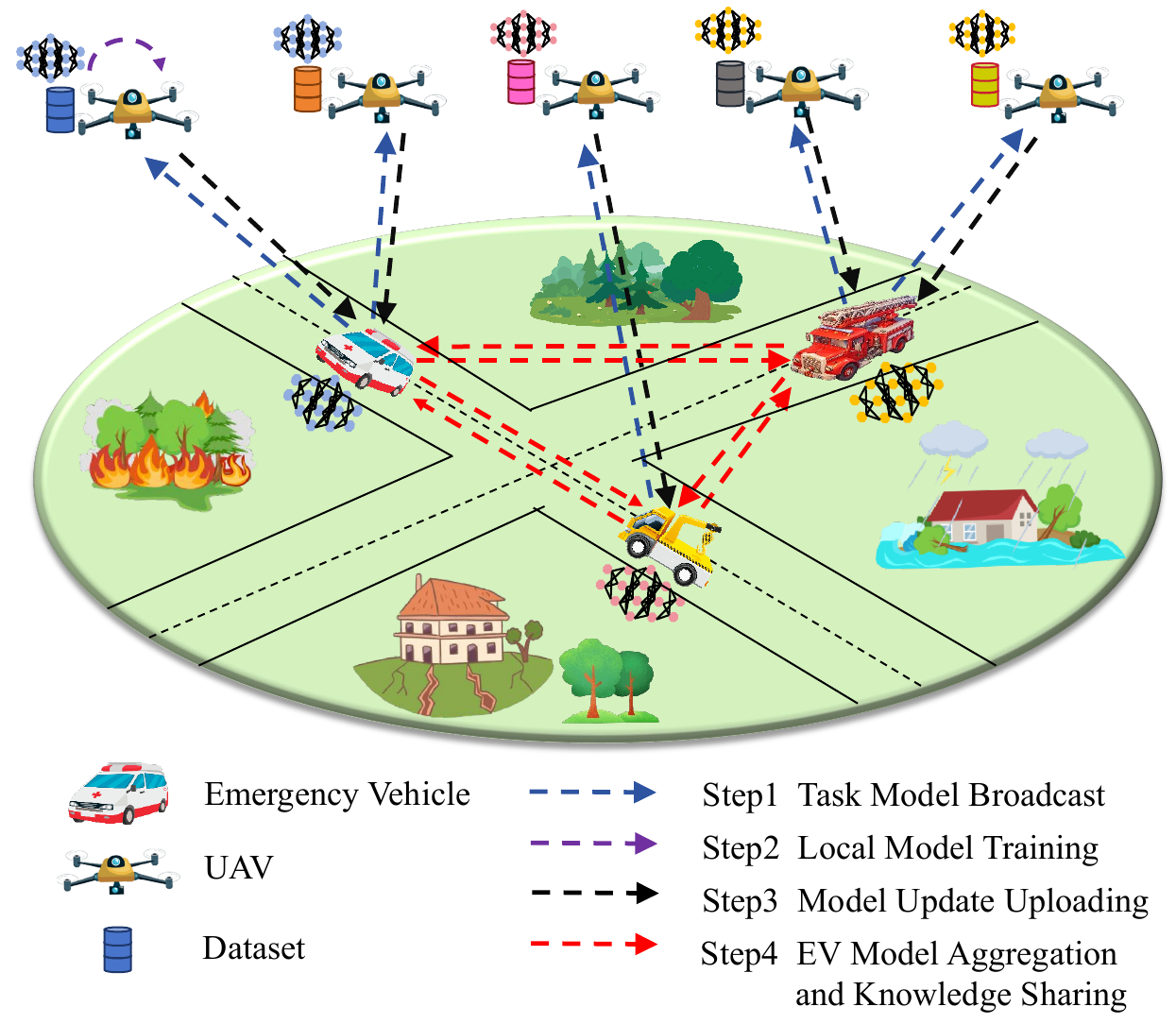}  
    \caption{UAV swarm assisted multi-task FL scenario}
    \label{fig:shiyitu}
\end{figure}

As shown in Fig.\hyperref[fig1]{1}, the considered system consists of $M$ Emergency Vehicles (EVs) indexed by $\mathcal{M}=\{1,2, \cdots, M\}$ and $N$ Unmaned Aerial Vehicles (UAVs) indexed by $\mathcal{N}=\{1,2, \cdots, N\}$. UAVs are deployed in different areas and altitudes to collect data for training models across multiple tasks. To ensure the robustness of the network, each EV $m$ is responsible for coordinating a different task $m$ respectively, and the EVs can communicate with each other for information exchange. Each UAV \(n\) has a local dataset \(\mathcal{D}_n\) containing \(D_n = \left|\mathcal{D}_n\right|\) data samples, which can be used for \(M\) tasks.
The entire dataset is
denoted by $\mathcal{D}=\cup\left\{\mathcal{D}_n\right\}_{n=1}^N$ with the total number of samples $D=\sum_{n=1}^N D_n$. Given a data sample $(\boldsymbol{x}, \boldsymbol{y}) \in \mathcal{D}$ collected by UAV, $\boldsymbol{x} \in \mathbb{R}^d$
is the $d$-dimensional input data vector, $\boldsymbol{y}=[y_1,y_2,\cdots,y_M] \in \mathbb{R}^M$ is the corresponding $M$-dimensional ground-truth label and each dimension corresponds to a specific task.  The neural network model for each task \(m\) can be divided into an image feature extractor \(w_m^s\) and a predictor \(w_m^u\). Since the image dataset collected by UAVs is used to perform multiple tasks, there may be potential correlations in the lower-level image features required by different tasks. This enables us to explore the related knowledge of cross-task feature extractors to facilitate knowledge sharing, thereby improving the performance of these tasks. It is important to note that, in order to maintain model performance, the predictors $w_m^u$ must be unique for each task $m$. Let $f_{m,n}(\boldsymbol{x},\boldsymbol{y}; \boldsymbol{w}_{m,n})$ denote the sample-wise loss function for task $m$ on UAV $n$, $\boldsymbol{w}_{m,n}$ can be split into two parts $[\boldsymbol{w}_{m,n}^s,\boldsymbol{w}_{m,n}^u]$, where $\boldsymbol{w}_{m,n}^s$ represents the feature extractor and $\boldsymbol{w}_{m,n}^u$ represents the predictor layer of UAV $n$ for task $m$. Thus, the local loss function that UAV \(n\) uses to measure the model error for task \(m\) is defined as:
\begin{equation}
     {F_{m,n}(\boldsymbol{w}_{m,n})=\frac{1}{D_{n}}\sum_{(\boldsymbol{x},\boldsymbol{y})\in\mathcal{D}_{n}}f_{m,n}(\boldsymbol{x},\boldsymbol{y};\boldsymbol{w}_{m,n})}
\end{equation}
Therefore, the global loss function for task $m$ is given by
\begin{equation}
F_m(\boldsymbol{w}_m)=\sum\nolimits_{n=1}^N \frac{D_n}{D}  F_{m,n}(\boldsymbol{w}_{m})
\end{equation}

Although models for different tasks share the same low-level feature extractor architecture, the high-level feature extractors may vary across tasks, and subsequent predictors can also adopt different architectures. As a result, the model parameters $\boldsymbol{w}_m,\forall m\in\mathcal{M} $ for different tasks can have different dimensions. We use $\boldsymbol{w}=[\boldsymbol{w}_1,\boldsymbol{w}_2,\cdots\boldsymbol{w}_M]$ to represent the set of all task parameters, and $\mathcal{F}(\boldsymbol{w})=[F_1(\boldsymbol{w}_1),F_2(\boldsymbol{w}_2),\cdots,F_M(\boldsymbol{w}_M)]$ to represent the loss function of all tasks. For simplicity, we omit the subscript \(m\) from \(F\) and use \(F_{n}(\boldsymbol{x}, \boldsymbol{y}; \boldsymbol{w}_{m,n})\) to denote the loss function of task \(m\) on UAV \(n\). It should be noted that due to resource and energy constraints, we assume that each UAV can only participate in one task per round. To optimize the network’s average performance across all tasks, the objective function is defined as
\begin{equation}
    F(\boldsymbol{w})=\frac{1}{M} \sum\nolimits_{m=1}^M  F_{m}(\boldsymbol{w}_{m})
\end{equation}

\begin{figure}[h]\label{fig2}
    \centering
    \includegraphics[width=0.49\textwidth]{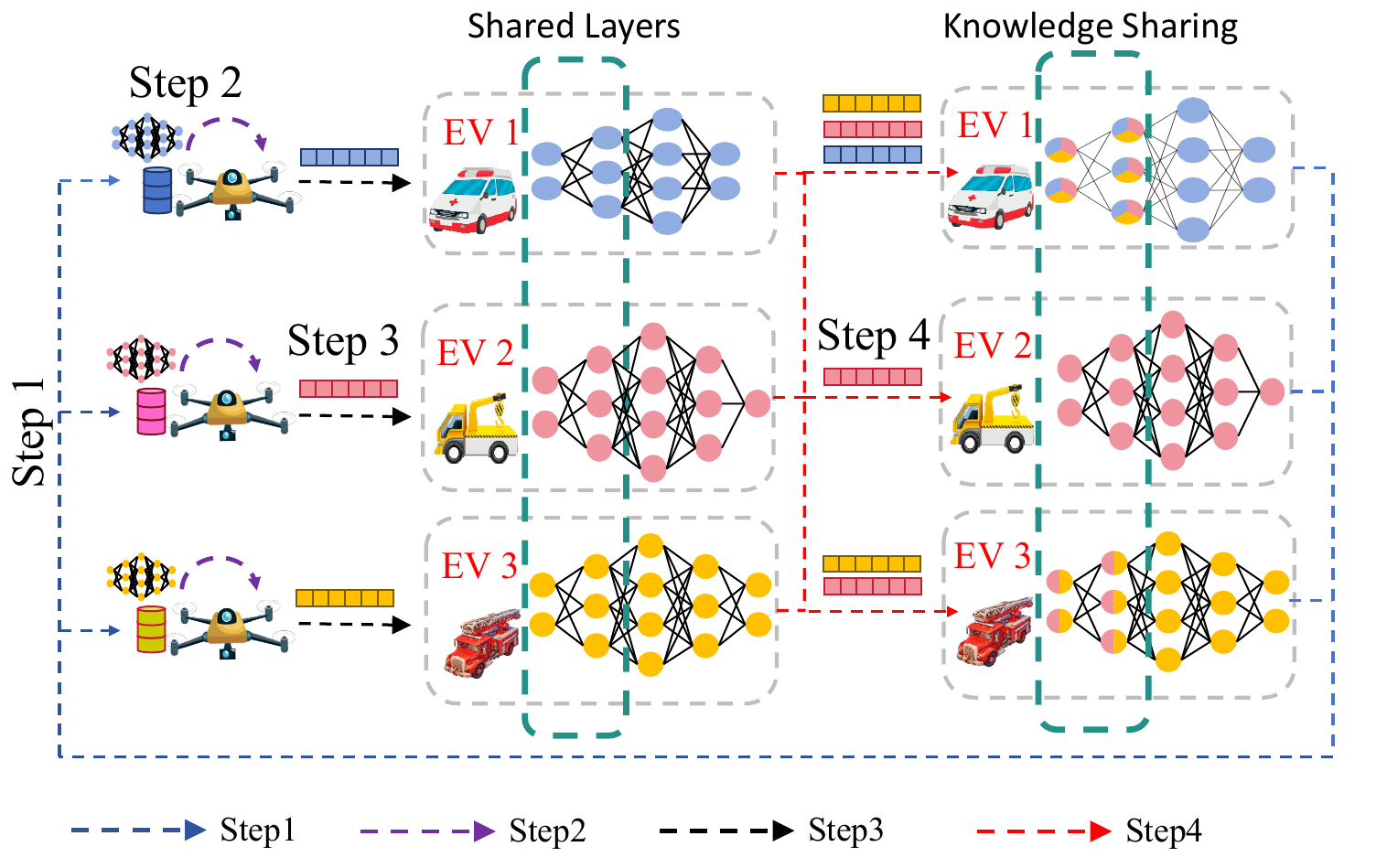}  
    \caption{The proposed scheme with task knowledge sharing, where EV1 shares knowledge with both EV2 and EV3, while EV3 shares knowledge with EV2.
 }
    \label{knowlege}
\end{figure}

To achieve the objectives outlined in (3), the training steps of the proposed approach are depicted in Fig. \hyperref[fig2]{2} and detailed as follows.

$\bullet$ \textbf{(Step 1 UAV-EV Association and Task Model Broadcast)} At the beginning of each round, each UAV is assigned to an EV via the UAV-EV association algorithm, which will be detailed in Section \hyperref[section4]{IV}, and participates in the corresponding training task. Each EV broadcasts its latest model parameters $\boldsymbol{w}_{m,t}$ for task $m$ to its associated UAVs. Let $\beta_{m,n}^t \in \{0,1\}$ denote the association indicator, where $\beta_{m,n}^t = 1$ indicates that UAV $n$ is associated with EV $m$ in round $t$, and $\beta_{m,n}^t = 0$ otherwise. We define $\boldsymbol{\beta_{m}^t} = [\beta_{m,1}^t, \beta_{m,2}^t, \cdots, \beta_{m,N}^t]^T \in \mathbb{R}^N$ to represent the UAV association for task $m$ in round $t$. Similarly, $\boldsymbol{\beta_t} = [\boldsymbol{\beta_{1}^t}; \boldsymbol{\beta_{2}^t}; \cdots; \boldsymbol{\beta_{M}^t}]^T$ represents the overall UAV-EV association across all tasks.

$\bullet$ \textbf{(Step 2 Local Model Training)} Upon receiving the latest corresponding task-specific global model $\boldsymbol{w}_{m,t}$ from the EV, each UAV $n$ initialize $\boldsymbol{w}_{m,n,t}=\boldsymbol{w}_{m,t}$ and updates the feature extraction layer $\boldsymbol{w}_{m,n,t}^{s}$ and task-specific layer $\boldsymbol{w}_{m,n,t}^{u}$ respectively.

\noindent$\boldsymbol{w}_{m,n,t}^{s}$ is updated as
\begin{equation}
   \boldsymbol{w}_{m,n,t+1}^{s}=\boldsymbol{w}_{m,n,t}^{s}-\eta\nabla_{s}F_{n}(\boldsymbol{w}_{m,n,t}^{s},\boldsymbol{w}_{m,n,t}^{u})
\end{equation}
$\boldsymbol{w}_{m,n,t}^{u}$ is updated as 
\begin{equation}
   \boldsymbol{w}_{m,n,t+1}^{u}=\boldsymbol{w}_{m,n,t}^{u}-\eta\nabla_{u}F_{n}(\boldsymbol{w}_{m,n,t}^{s},\boldsymbol{w}_{m,n,t}^{u})
\end{equation}
where $\eta>0$ is the learning rate .


$\bullet$ \textbf{(Step 3 Model Update $\&$ Parameter Uploading )} Once local model training completed, each UAV $n$ uploads its gradient update $\boldsymbol{G}_{m,n,t}$ to its associated EV.
$\boldsymbol{G}_{m,n,t}$ is given by
\begin{equation}
\boldsymbol{G}_{m,n,t}=\frac{1}{\eta}\left(\boldsymbol{w}_{m,n,t+1}-\boldsymbol{w}_{m,n,t}\right)
\end{equation}
Note that the gradient can be expressed as $\boldsymbol{G}_{m,n,t}=[\boldsymbol{G}_{m,n,t}^{s},\boldsymbol{G}_{m,n,t}^{u}]$, where $\boldsymbol{G}_{m,n,t}^{s}$ and $\boldsymbol{G}_{m,n,t}^{u}$ represent the feature extractor gradient and task-specific layer gradient respectively.

$\bullet$ \textbf{(Step 4 EV Model Aggregation and Knowledge Sharing)}
After receiving the gradients from all associated UAVs, each EV updates its predictor as
\begin{equation}
\boldsymbol{w}_{m,t+1}^{u}=\boldsymbol{w}_{m,t}^{u}-\eta\frac{\sum\nolimits_{n=1}^{N}\beta_{m,n}^{t}D_{n}G^u_{m,n,t}}{\sum\nolimits_{n=1}^{N}\beta_{m,n}^{t}D_{n}}
\end{equation}
Then each EV transmits $\boldsymbol{G}^s_{m,t}=\frac{\sum\nolimits_{n=1}^{N}\beta_{m,n}^{t}D_{n}G^s_{m,n,t}}{\sum\nolimits_{n=1}^{N}\beta_{m,n}^{t}D_{n}}$ to other EVs for potential knowledge sharing. Based on the task affinity indicator (introduced in Section \hyperref[section4]{IV}), each EV $m$ updates its feature extractor using the gradients from task $m$ and its related tasks as follows:
\begin{equation}
\boldsymbol{w}_{m,t+1}^{s}=\boldsymbol{w}_{m,t}^{s}-\eta\frac{{\textstyle \sum\nolimits _{i\in S_{m,t}}}D_{i,t}\boldsymbol{G}^s_{i,t} }{\sum\nolimits_{i\in S_{m,t}}D_{i,t}} 
\end{equation}
where \(S_{m,t}\) denotes the set of EVs involved in the feature extractor update for task \(m\) in round \(t\). This implies that the gradients of the feature extractors from the EVs in \(S_{m,t}\) contribute positively to the learning of task \(m\). Additionally, $D_{i,t} = \sum\nolimits_{n=1}^{N} \beta_{i,n}^t D_n$ denotes the total number of training samples from UAVs associated with EV $i$ in round $t$.


Steps 1 to 4 are repeated until the predetermined number of iterations is reached.
\subsection{Computation Model}
As mentioned earlier, the model architectures for different tasks may vary, leading to different parameter dimensions and computational resource requirements for each task. Let $C_{m,n}$ denote the CPU cycles required to process one data sample when UAV $n$ trains model of task $m$. With dynamic frequency adjustment technology, each UAV's CPU frequency can be adjusted within the range $0 < f_n < f_{max}$. Let $f_{n,t}$ denote the computation frequency (CPU cycles per second) of UAV $n$ in round $t$. The local training time of UAV $n$ in round $t$ is then given by:
\begin{equation}
T_{n,t}^{comp}=\frac{KD_n C_{n,t}}{f_{n,t}}
\end{equation}
where $C_{n,t}= \sum\nolimits _{m=1}^{M} \beta_{m,n}^tD_{n}$, $K$ is the number of local iterations and the computational energy consumption is
\begin{equation}\label{eq10}
E_{n,t}^{comp}=\varsigma KD_n C_{n,t}f_{n,t}^2
\end{equation}
where $\varsigma$ is the energy coefficient depending on the chip architecture. The computational cost of global model aggregation on an EV is minimal compared to model training and parameter transmission. Therefore, the associated energy consumption and time delay are negligible due to the EV's superior computing capacity.
\subsection{Communication Model}
The air-to-ground communication link is the probabilistic superposition of the line-of-sight(LOS) and NLOS channels\cite{b37}. The path loss between UAV $n$ and EV $m$ is
\begin{equation}
    L_{m,n,t}=\begin{cases}\alpha_0(d_t^{m,n})^{-\nu},&\text{if LoS link,}\\\mu^\text{NLoS}{\alpha_0}(d_t^{m,n})^{-\nu},&\text{if NLoS link,}\end{cases}
\end{equation}
where $\alpha_0$ represents the channel power gain when the reference distance is 1m and $\nu$ represents the path loss exponent. $\mu^{\text{NLoS}}$ is the additional attenuation coefficient of the NLoS link and $d_t^{m,n}=\sqrt{(x_{u,t}-x_{k,t})^{2}+(y_{u,t}-y_{k,t})^{2}+H^{2}}$ is the distance. The probability of existing an LOS channel is calculated as
$P_{m,n,t}^{\mathrm{LoS}}=\frac1{1+a\exp{(-b(\theta_{m,n,t}-a))}}$, where $a$ and $b$ are the coefficients related to the environment, $\theta_{m,n,t}=\frac{180}{\pi}\sin^{-1}(\frac{H}{d_t^{m,n}})$ is the elevation angle in degree. Accordingly, the probability of NLoS can be given by $P_{m,n,t}^{\mathrm{NLoS}}=1-P_{m,n,t}^{\mathrm{LoS}}$. Thus, the channel gain is given by
\begin{equation}
h_t^{m,n}=\left(P_{m,n,t}^{\mathrm{LoS}}+\mu^{\mathrm{NLoS}}P_{m,n,t}^{\mathrm{NLoS}}\right)\alpha_0(d_t^{m,n})^{-\nu}
\end{equation}
We adopt frequency division multiple access (FDMA) for communication, with the total bandwidth denoted by \( B \). In each communication round, UAV \( n \) is allocated a proportion \( \gamma_{n,t} \) of the uplink bandwidth for gradient uploading. Let \( \boldsymbol{\gamma}_t = \left( \gamma_{1,t}, \gamma_{2,t}, \dots, \gamma_{N,t} \right) \) represent the resource allocation vector in round \( t \), and \( p_{n,t} \) the transmit power of UAV \( n \), constrained by the maximum limit \( p_{n, \text{max}} \). The achievable transmit rate between UAV \( n \) and EV \( m \) is expressed as:
\begin{equation}
r_{m,n,t} = \gamma_{n,t} B \log_2\left(1+\frac{p_{n,t} h_{m,n,t}}{\gamma_{n,t} B N_0}\right)
\end{equation}
where \( N_0 \) is the noise power spectral density. Let \( Z_m \) denote the gradient size for task \( m \). If UAV \( n \) is scheduled to train task \( m \), its transmission time is \( T_{m,n,t}^{\mathrm{comm}} = \frac{Z_m}{r_{m,n,t}} \). Then the actual transmission time of UAV \( n \) is $T_{n,t}^{\mathrm{comm}} = \sum_{m=1}^{M} \beta_{m,n}^t T_{m,n,t}^{\mathrm{comm}}$
and the corresponding transmission energy consumption is $E_{n,t}^{\mathrm{comm}} = p_{n,t} T_{n,t}^{\mathrm{comm}}$.
Similar to \cite{b20}, the energy consumption for maintaining the UAV's flight is neglected. Therefore, the total energy consumed by UAV $n$ in round $t$ is 
$E_{n,t}=E_{n,t}^{\mathrm{comp}}+E_{n,t}^{\mathrm{comm}}$.

\section{Problem Analysis And Problem Formulation}
In this section, we thoroughly analyze the critical challenges in our scenario, and then formulate the specific problem to be addressed.
\subsection{Problem Analysis}
In UAV swarm based multi-task emergency rescue scenarios, after the task training completed, UAVs are dispatched to various areas to perform specific tasks. Therefore, task completion performance and timeliness are crucial considerations. The primary objective is to capture the correlations among tasks to promote knowledge sharing and effectively meet the heterogeneous requirements of multiple tasks under resource constraints. As training advances and the environment evolves, both task requirements and the resource demands for UAVs to participate in various tasks exhibit dynamic and time-varying characteristics. Balancing UAV energy consumption while effectively capturing and satisfying these dynamic task demands poses a significant challenge. Furthermore, since multiple tasks are based on the same dataset, potential correlations among tasks can exist. These correlations, however, are not static and evolve dynamically during training, making it crucial to capture and utilize them to enable knowledge sharing across related tasks. 

This work therefore aims to optimize the dynamic UAV-EV association under energy and resource constraints, while leveraging time-varying task correlations to facilitate knowledge sharing, thereby maximizing the average performance across all tasks. However, due to the complex evolution of machine learning models during the learning process, the factors influencing the network's performance remain unknown, we first conduct a convergence analysis of the proposed scheme to identify the factors affecting task performance, and subsequently formulate the problem to be addressed.

\subsection{Convergence Analysis}
Before conducting the convergence analysis, we first present several necessary assumptions.

\noindent$\textbf{\textit{Assumption 1}}$. The expected squared norm of the local feature extractor gradient and predictor gradient for each task \(m\) and UAV \(n\) is uniformly bounded.
\begin{equation}
\mathbb{E}[\left\|\nabla_{s}F_{m,n}(\boldsymbol{w}_{m,n,t})\right\|^2]\leq \epsilon_s  ^2,
\text{ } \mathbb{E}[\left\|\nabla_{u}F_{m,n}(\boldsymbol{w}_{m,n,t})\right\|^2]\leq \epsilon_u ^2.
\end{equation}
\noindent$\textbf{\textit{Assumption 2}}$. For each task $m$ and UAV $n$, there exist $L_s,L_u,L_{su}$ and $L_{us}$ such that:

\noindent$\bullet$$\text{  }\nabla_s F_{m,n}(\boldsymbol{w}_{m}^{s},\boldsymbol{w}_{m}^{u})$ is $L_s$-Lipschitz continuous with $\boldsymbol{w}_{m}^{s}$ and $L_{su}$-Lipschitz continuous with $\boldsymbol{w}_{m}^{u}$:
\begin{equation}
    ||\nabla_s F_{m,n}(\bar{\boldsymbol{w}}^{s}, \boldsymbol{w}_{m}^{u})-\nabla_s F_{m,n}(\check{\boldsymbol{w}}^{s},\boldsymbol{w}_{m}^{u})||\le L_s||\bar{\boldsymbol{w}}^{s}-\check{\boldsymbol{w}}^{s}||
\end{equation}
\begin{equation}
    ||\nabla_s F_{m,n}(\boldsymbol{w}_{m}^{s}, \bar{\boldsymbol{w}}^{u})-\nabla_s F_{m,n}(\boldsymbol{w}_{m}^{s},\check{\boldsymbol{w}}^{u})||\le L_{su}||\bar{\boldsymbol{w}}^{u}-\check{\boldsymbol{w}}^{u}||
\end{equation}

\noindent$\bullet$$\text{  }\nabla_u F_{m,n}(\boldsymbol{w}_{m}^{s},\boldsymbol{w}_{m}^{u})$ is $L_u$-Lipschitz continuous with respect to $\boldsymbol{w}_{m}^{u}$ and $L_{us}$-Lipschitz continuous with respect to $\boldsymbol{w}_{m}^{s}$, the details are omitted here for simplicity.

\noindent$\textbf{\textit{Assumption 3}}$. For correlated tasks \(i\) and \(j\), the gradient difference of the feature extractors based on the same parameter is bounded.
\begin{equation}
    ||\nabla_s F_{i,n}(\boldsymbol{w}^{s}, \boldsymbol{w}_{i}^{u})-\nabla_s F_{j,n}(\boldsymbol{w}^{s},\boldsymbol{w}_{j}^{u})||\le \delta^2
\end{equation}

Assumption 1-2 are commonly used in the convergence analysis of federated learning \cite{b15} \cite{b39} \cite{b40}. Assumption 2 implies that when the input variation of the network is limited, the output variation is also constrained within a certain range, which is reasonable as most deep neural networks satisfy this condition. Assumption 3 states that when tasks are correlated, the underlying image features required are similar. Thus, when the same UAV computes feature extractor parameters for different tasks using the same parameters, the gradient directions should be similar \cite{Caruana}, implying that the gradient difference is bounded. Based on these assumptions, we have the following lemma.

\begin{Lemma}\label{lemma1}
For any task $m,\forall m\in\mathcal{M}$, the difference between its loss functions in two consecutive rounds satisfies
\begin{equation}
\begin{aligned}
&F_{m}(\boldsymbol{w}_{m,t+1})-F_{m}(\boldsymbol{w}_{m,t})\\
&\leq\left\langle\nabla_{s}F_{m}(\boldsymbol{w}_{t}^{s},\boldsymbol{w}_{m,t+1}^{u}),\boldsymbol{w}_{t+1}^{s}-\tilde{\boldsymbol{w}}_{m,t+1}^{s}\right\rangle\\
&\hspace{0.5cm}+\left\langle\nabla_{s}F_{m}(\boldsymbol{w}_{t}^{s},\boldsymbol{w}_{m,t+1}^{u}),\tilde{\boldsymbol{w}}_{m,t+1}^{s}-\boldsymbol{w}_{t}^{s}\right\rangle\\
&\hspace{0.5cm}+\left\langle\nabla_{u}F_{m}(\boldsymbol{w}_{t}^{s},\boldsymbol{w}_{m,t}^{u}),\boldsymbol{w}_{m,t+1}^{u}-\boldsymbol{w}_{m,t}^{u}\right\rangle\\
&\hspace{0.5cm}+\frac{L_{s}}{2}\left\|\boldsymbol{w}_{t+1}^{s}-\boldsymbol{w}_{t}^{s}\right\|^{2}+\frac{L_{u}}{2}\left\|\boldsymbol{w}_{m,t+1}^{u}-\boldsymbol{w}_{m,t}^{u}\right\|^{2}
\end{aligned}
\end{equation}
\end{Lemma}
where $\tilde{\boldsymbol{w}}_{m,t+1}^{s}$ represents the feature extractor parameters at round \( t+1 \) obtained by updating task \( m \) using only its own gradient. The proof can be found in Appendix \hyperref[AppendixA]{A} in the supplementary material. Based on lemma \hyperref[lemma1]{1}, we have the following theorem
\begin{theorem}\label{theorem1}
Given the UAV-EV association $\beta_t$ in round $t$, when the learning rate $\eta < \min\left(\frac{1}{2(2L_{s} + L_{su})},\frac{1}{2(L_{u} + L_{su})}\right)$, the difference of the loss function for task $m$ between two consecutive rounds is bounded by:
\begin{equation}\label{eq19}
\begin{aligned}
&F_{m}(\boldsymbol{w}_{m,t+1})-F_{m}(\boldsymbol{w}_{m,t})\\
&\leq (-\frac{\eta }{2}+L\eta ^2) \left [  \mathbb{E}\left \| \nabla_{s}F_{m}(\boldsymbol{w}_{m,t})\right \|^2+ \mathbb{E}\left \| \nabla_{u}F_{m}(\boldsymbol{w}_{m,t})\right \|^2\right ]\\
&\hspace{0.5cm} +8(1+2L_{su})(1\!-\!\frac{\sum_{i\in S_{m,t}}D_{i,t}}{D} )^2\epsilon_s^2\\
&\hspace{0.5cm}+4(1+2L_{su})(1-\frac{D_{m,t}}{\sum_{i\in S_{m,t}}D_{i,t}})^2\delta^2+\Omega_t
\end{aligned}
\end{equation}
\end{theorem}
\noindent where $L=\max \{2L_s+L_{su},L_u+L_{su}\}$ and $ \Omega_t=(5+8L_{su}+\eta(2\eta L_s+\eta L_{su}+\frac{1}{2} ))(1-\frac{D_{m,t}}{D} )^2\epsilon _s^2+(\frac{\eta}{2}+\eta^2(L_u+L_{su}) )(1-\frac{D_{m,t}}{D} )^2\epsilon_u^2$.

\noindent \textit{Proof}: See Appendix \hyperref[AppendixB]{B} in the supplementary material.

\noindent \textbf{Remark 1}\label{remark1}: According to Theorem \hyperref[theorem1]{1}, the reduction in the loss of task \( m \) depends on the number of samples used for task \( m \) and the degree of similarity in the required underlying features between task \( m \) and the other tasks in \( S_{m,t} \). The second term \( \left(1 - \frac{\sum_{i \in S_{m,t}} D_{i,t}}{D} \right)^2 \epsilon_s^2 \) reflects the impact of the data volume used to update task \( m \), decreasing as more data is allocated to \( m \). The term \( \left(1 - \frac{D_{m,t}}{\sum_{i \in S_{m,t}} D_{i,t}} \right)^2 \delta^2 \) represents the impact of the difference in required features between task \( m \) and the tasks in \( S_{m,t} \). A smaller \(\delta\) indicates more similar underlying features. This term becomes zero when only the gradients of task \( m \) are used. Incorporating gradients from related tasks increases \( \sum_{i \in S_{m,t}} D_{i,t} \), reducing the second term but potentially increasing the third due to feature differences. Thus, accurately leveraging gradients from related tasks is crucial for minimizing the right-hand side of \hyperref[eq19]{(19)}, thereby improving task performance and accelerating convergence.

\noindent \textbf{Remark 2}\label{remark2}: Due to the dynamic UAV-EV association and variations in task training difficulty, the training progress of tasks differs and task demands exhibit time-varying characteristics. Given the diminishing marginal returns in neural network training, more attention should be attached toward tasks with slower progress. In addition, the ability of each task to improve related tasks' performance through knowledge sharing varies. Thus, dynamic weights should be assigned to each task across different rounds.

Based on the above analysis, problem can be reformulated as follows:
\begin{subequations}\label{p1}
\renewcommand{\theequation}{20\alph{equation}}
\begin{align}
\mathcal{P}_1: &\max_{\left\{\boldsymbol{\beta_t},\boldsymbol{\gamma_t},\boldsymbol{p_t},\boldsymbol{f_t}\right\}_{t=1}^{T}}  \sum_{t=1}^{T}\sum_{m=1}^{M} \alpha_m^t\sum_{n=1}^{N}\beta _{m,n}^t D_n \tag{20}\\
\text{s. t. } & \sum\nolimits_{t=1}^{T} (E_{n,t}^{\text{comp}} + E_{n,t}^{\text{comm}}) \leq E_{n,\text{max}}, \quad \forall n \in \mathcal{N}, & \tag{20a} \label{20a} \\
&T_{n,t}^{\text{comp}} + T_{n,t}^{\text{comm}} \leq T_{\text{max}}, \quad \forall n \in \mathcal{N}, \forall t, \tag{20b} \label{20b}\\
& \beta_{m,n}^t \in \{0,1\}, \sum\nolimits_{m=1}^M \beta_{m,n}^t = 1, \quad \forall m ,  n , t, & \tag{20c} \label{20c} \\
& \sum\nolimits_{n=1}^N \beta_{m,n}^t \geq \delta_m, \quad \forall m \in \mathcal{M}, \forall t, & \tag{20d} \label{20d} \\
& \gamma_{n,t}>0, \sum\nolimits_{n=1}^{N} \gamma_{n,t}= 1, \quad \forall n \in \mathcal{N}, \forall t, & \tag{20e} \label{20e}  \\
& 0 \leq p_{n,t} \leq p_{n,\text{max}}, \quad \forall n \in \mathcal{N}, \forall t, & \tag{20f} \label{20f}\\
&0 < f_{n,t} \leq f_{n,max}, \quad \forall n \in \mathcal{N},\forall t  \tag{20g}\label{20g}\\
&S_{m,t}\subseteq \mathcal{M} ,\quad \forall m \in \mathcal{M},\forall t\tag{20h}\label{20h}. 
\end{align}
\end{subequations}
Constraint \hyperref[20a]{20(a)} ensures that the total energy consumption of each UAV does not exceed its energy capacity. \hyperref[20b]{20(b)} enforces the time limit for each training round, while \hyperref[20c]{20(c)-20(d)} define the UAV-EV associations in each round. Specifically, each UAV can only be associated with one EV, and the number of UAVs associated with each EV must meet a minimum threshold to ensure the performance of each task. \hyperref[20e]{20(e)} regulate the allocation of bandwidth. and \hyperref[20f]{20(f)} and \hyperref[20g]{20(g)} limit the transmission power and computing frequency of each UAV, respectively. \hyperref[20h]{20(h)} represents the EV group used for feature extractor update of task $m$ in round $t$.

The parameter \( \alpha_m^t \) in \hyperref[p1]{(20)} measures the importance weight of task \( m \) in round \( t \). Directly solving problem \( \mathcal{P}_1 \) requires knowledge of the channel state \( h_t \), task importance weights \( \alpha_m^t \), and the energy states of all UAVs across all rounds \( t \), which is impractical in real-world scenarios. To eliminate non-causal information and enable online dynamic UAV-EV association using only current-round information while satisfying constraint \hyperref[20a]{(20a)}, we employ Lyapunov theory\cite{b48} to construct virtual queue \( Q_{n,t} \) for each UAV \( n \). This virtual queue represents the gap between the cumulative energy consumption and the budget up to round \(t\), evolving according to the following formula:
\begin{equation}
Q_{n, t+1}\!=\!\max \left\{Q_{n,t}+{E}_{n, t}^{comp}+{E}_{n, t}^{comm}-\frac{E_{max}}{T}, 0\right\}
\end{equation}\label{eq21}

To satisfy constraint \hyperref[20a]{(20a)}, it is crucial to maintain the stability of $Q_{n,t},\forall n$, and the Lyapunov function\cite{b47} is defined as follows to measure the congestion level of the queues:
\begin{equation}
L\left(\boldsymbol{Q}^{t}\right)=\frac{1}{2} \sum_{n=1}^NQ_{n,t}^{2},
  \end{equation}
where $\boldsymbol{Q}^{t}=\left(Q_{1,t}, \ldots, Q_{N,t}\right)$ is the vector of virtual queues. The Lyapunov drift function conditioned on $\boldsymbol{Q}^{t}$ is defined as
\begin{equation}
\Delta\left(\boldsymbol{Q}^{t}\right)=\mathbb{E}\left[L\left(\boldsymbol{Q}^{t+1}\right)-L\left(\boldsymbol{Q}^{t}\right) \mid \boldsymbol{Q}^{t}\right]
\end{equation}

According to Lyapunov optimization theory, the increase of the Lyapunov function should be constrained to stabilize $\boldsymbol{Q}^{t}$. We first derive the upper bound of $\Delta\left(\boldsymbol{Q}^{t}\right)$, as shown in the following lemma:
\begin{Lemma}\label{lemma2}
The Lyapunov drift function in round $t$ is bounded by
\begin{equation}
\Delta\left(\boldsymbol{Q}^{t}\right)\leq B + \sum\nolimits_{n=1}^N Q_{n,t} \left(E_{n,t}^{\text{comp}}\!+\! E_{n,t}^{\text{comm}} - \frac{E_{\text{n,max}}}{T} \right)
\end{equation}
where $B=\frac{N}{2}\max (\bar{E}_n^2,\Xi )$, $\bar{E}_n=\frac{E_{n,max}}{T}$ and $\Xi=\max_n \left (p_{n}^{max}T_{max}+\max_m \varsigma K L_b C_{m,n}\left(f_n^{max}\right)^2\right )$.
\end{Lemma}

\noindent$\mathit{proof:}$
\begin{equation}
\begin{aligned}
\Delta\left(\boldsymbol{Q}^{t}\right)&=L\left(\boldsymbol{Q}^{t+1}\right) - L\left(\boldsymbol{Q}^{t}\right) \\
&= \frac{1}{2} \sum\nolimits_{n=1}^N \left(Q_{n,t+1}^2 - Q_{n,t}^2\right) \\
&\overset{(a)}{\leq} \frac{1}{2} \sum\nolimits_{n=1}^N \left(E_{n,t}^{\text{comp}} + E_{n,t}^{\text{comm}} - \bar{E}_n  \right)^2 \\
&\quad + \sum\nolimits_{n=1}^N Q_{n,t} \left(E_{n,t}^{\text{comp}} + E_{n,t}^{\text{comm}} - \bar{E}_n \right) \\
&\overset{(b)}{\leq} B + \sum\nolimits_{n=1}^N Q_{n,t} \left(E_{n,t}^{\text{comp}} + E_{n,t}^{\text{comm}} - \bar{E}_n \right)
\end{aligned}
\end{equation}
(a) comes from \hyperref[eq21]{(21)}, (b) follows from $E_{n,t}^{\text{comm}} \leq p_{n,max}T_{max}$ and $E_{n,t}^{\text{comp}}\leq \max_m \varsigma    K L_b C_{m,n}\left(f_n^{max}\right)^2$, so the energy consumption is constrained, then the proof is completed.

Denote $U^t=\sum_{m=1}^{M} \alpha_m^t\sum_{n=1}^{N}\beta _{m,n}^t D_n$, the upper bound of the Lyapunov drift-plus-penalty metric is given by
\begin{equation}\label{eq26}
\begin{aligned}
\Delta(\boldsymbol{Q}^{t})&-VU^t=\Delta(\boldsymbol{Q}^{t})-V \sum\nolimits_{m = 1}^{M} \alpha_m^t\sum\nolimits_{n = 1}^{N}\beta _{m,n}^t D_n \\
&\leq B+ \sum\nolimits_{n=1}^N Q_{n,t} \left(E_{n,t}^{\text{comp}} + E_{n,t}^{\text{comm}} - \frac{E_{\text{n,max}}}{T} \right) \\
&\hspace{3em} -V\sum\nolimits_{m = 1}^{M} \alpha_m^t\sum\nolimits_{n = 1}^{N}\beta _{m,n}^t D_n 
\end{aligned}
\end{equation}

The parameter \( V \) in \hyperref[eq26]{(26)} is introduced to balance queue stability and training performance improvement, with a larger \( V \) placing greater emphasis on enhancing learning performance. System stability is ensured by limiting the growth of \( Q_{n,t}\). To achieve this, we aim to minimize the drift plus penalty term, simultaneously stabilizing the system and improving the task performance. After omitting the constant term, the problem can be reformulated as:
\begin{subequations}
\renewcommand{\theequation}{\alph{equation}}\label{eq27}
\begin{align}
\mathcal{P}_2: &\min_{\boldsymbol{\beta_{t}},\boldsymbol{\gamma_{t}}, \boldsymbol{p_{t}},\boldsymbol{f_{t}}}\sum_{n=1}^N Q_{n,t} E_{n,t} 
- V \sum_{m = 1}^{M} \alpha_m^t \sum_{n = 1}^{N} \beta_{m,n}^t D_n 
\tag{27}\\
\text{s. t. } & \quad \hyperref[p1]{(20b)-(20h)}  \quad \tag{27a}
\end{align}
\end{subequations}

Problem $\mathcal{P}_2$ remains difficult to solve, as it involves both integer and continuous variables and is therefore a mixed integer nonlinear programming(MINLP) problem. There is no method to solve it directly, and we will decompose problem $\mathcal{P}_2$ in section \hyperref[section4]{IV} to solve it.

\section{Proposed Algorithm}\label{section4}
So far, $\alpha_m^t$ in \hyperref[eq27]{(27)} and $S_{m,t}$ in \hyperref[20h]{(20h)} are unclear. To solve problem $\mathcal{P}_2$, we first propose a task attention mechanism to measure the dynamic importance of tasks, i.e., $\alpha_m^t,\forall m\in\mathcal{M}$, and introduce a task affinity metric to capture time-varying task correlations which is essential for determining $S_{m,t},\forall m \in\mathcal{M}$, thereby facilitating knowledge sharing. Subsequently, we design the optimal resource allocation and UAV-EV association strategy.
\subsection{Task Attention Mechanism}
Task attention mechanism is introduced to dynamically determine the importance weight \( \alpha_m^t \) for each task in each round. According to Remark \hyperref[remark2]{2}, to ensure the overall performance improvement across multiple tasks, it is necessary to consider the differences in training progress between tasks and their contributions to enhancing the performance of other tasks, the details are as follows.
\subsubsection{Task Performance Balance}
In multi-task learning systems, significant performance differences between tasks can lead to prolonged delays and unnecessary resource waste. Given the diminishing marginal returns in neural network training, allocating more resources to nearly completed tasks yields minimal performance gains and wastes resources. Thus, dynamic task weights should account for each task's training progress. To address this, we monitor each task's performance over past training periods and adjust its training weight accordingly. Specifically, each EV \( m \) maintains a parameter \( \Gamma_{m,t} \) to record the current weighted cumulative performance of its task.
\begin{equation}
\Gamma_{m,t}=\varpi\Gamma_{m,t-1}+(1-\varpi)loss_{m,t-1}
\end{equation}
where \(loss_{m,t - 1}\) and \(\Gamma_{m,t - 1}\) represent the loss and the cumulative loss of task \(m\) in round $t-1$ respectively. \(\varpi\) is the parameter that balances the importance of historical cumulative loss and current loss. Then, the performance based weight of each task $m$ in round $t$ can be obtained as
\begin{equation}
\tilde{\Psi}_{m,t}=\frac{e^{\Gamma_{m,t}}}{\sum_{m=1}^Me^{\Gamma_{m,t}}}
\end{equation}

\subsubsection{Task Shapley Value}
The original purpose of Shapley values is to distribute rewards fairly and reasonably to each player in a cooperative game. In this paper, we propose the Task Shapley Value (TSV) to measure the marginal contribution of each task $m,\forall m\in \mathcal{M}$ to the performance improvement of all task models $m’,\forall m’\in \mathcal{M}$ in the system through task knowledge sharing. Specifically, we denote a subset of EVs as \(C\). The utility function \( v_m(C) \) is defined as the accuracy of task \( m \) obtained by applying the current feature extractor gradients of the EVs in \( C \) to the model of task \( m \). Subsequently, the Shapley value of task \( i \) can be computed as:
\begin{equation}
\phi_{i,t}=\frac{1}{Mn}\sum_{\substack{m=1}}^{M}\sum_{k=0}^{n-1}\sum_{\substack{|C|=k,\\\{i\}\notin C}}\frac1{\binom{n-1}k}(v_m(C\cup\{i\})-v_m(C)) 
\end{equation}
Then, the cumulative Shapley value of each task $m$ is updated as
\begin{equation}
\mathcal{I}_{m,t}=\kappa \mathcal{I}_{m,t-1}+(1-\kappa  )\phi_{m,t-1}
\end{equation}
A large $\mathcal{I}_{m,t}$ implies that task $M$ not only has its own performance improved, but also can enhance the performance of other related tasks through knowledge sharing. In order to promote knowledge sharing among tasks and improve the overall performance of multiple tasks, we should pay more attention to tasks with large $\mathcal{I}_{m,t}$, so the TSV-based weight can be calculated as
\begin{equation}
\hat{\Psi}_{m,t}=\frac{e^{\mathcal{I}_{m,t}}}{\sum_{m=1}^Me^{\mathcal{I}_{m,t}}}
\end{equation}
Finally, we can get the weight of each task in round $t$ based on the task attention mechanism as
\begin{equation}
\Psi_{m,t}=\frac{\tilde{\Psi}_{m,t}+\hat{\Psi}_{m,t}}{\sum_{m=1}^M(\tilde{\Psi}_{m,t}+\hat{\Psi}_{m,t})}
\end{equation}
\subsection{Task Affinity}
Although tasks share the same image dataset, inherent differences in task objectives and the dynamic nature of task requirements result in varying underlying feature needs, leading to time-varying task correlations. Simply merging feature extractors gradients across tasks can introduce conflicts, potentially degrading overall performance. Inspired by \cite{b38}, we introduce the task affinity index to dynamically track task correlations and mitigate negative transfer effects caused by conflicting tasks.
Specifically, the affinity of task $i$ to task $j$ in round $t$ is defined as 
\begin{equation}
\Theta_{i \to j}^t = 1 - \frac{L_j(\boldsymbol{w}_{j,t}^s, \boldsymbol{G}_{i,t}, \boldsymbol{G}_{j,t})}{L_j(\boldsymbol{w}_{j,t}^s, \boldsymbol{G}_{j,t})}
\end{equation}
where \( L_j(\boldsymbol{w}_{j,t}^s, \boldsymbol{G}_{j,t}) \) denotes the loss of task \( j \) on EV \( j \)'s test data after updating the feature extractor with \( \boldsymbol{G}_{j,t} \), and \( L_j(\boldsymbol{w}_{j,t}^s, \boldsymbol{G}_{i,t}, \boldsymbol{G}_{j,t}) \) denotes the loss after updating with both \( \boldsymbol{G}_{i,t} \) and \( \boldsymbol{G}_{j,t} \). Additionally, the cumulative task affinity is defined as $\Upsilon_{i \to j}^t = \ell \Upsilon_{i \to j}^{t-1} + (1-\ell) \Theta_{i \to j}^t$ and \( \ell \) is the balancing factor. As a result, \( S_{m,t} \) in \hyperref[20h]{(20h)} for each task $m$ only includes EVs corresponding to \(\Upsilon_{i \to m}^t > 0\), thus avoiding negative knowledge transfer caused by conflicting tasks.

So far, to solve problem $\mathcal{P}_2$, we have introduced a task attention mechanism to quantify the time-varying importance weight \( \alpha_m^t \) of each task and utilized task affinity to facilitate knowledge sharing among related tasks. To further solve problem $\mathcal{P}_2$, we focus on deriving the optimal resource allocation strategy and the optimal UAV-EV association.




\subsection{Frequency and Power Control}
After determining $\alpha_m^t,\forall m\in\mathcal{M}$ and $S_{m,t},\forall m\in\mathcal{M}$ in round $t$, we now focus on the UAV-EV association and resource allocation problems. These problems are divided into an inner problem and an outer problem: the inner problem addresses resource allocation given a specific UAV-EV association, while the outer problem determines the optimal UAV-EV association based on the solution to the inner problem.

For the inner problem, given \( \boldsymbol{\beta}_t \), problem $\mathcal{P}_2$ can be reformulated as:
\begin{subequations}\label{eq35}
\renewcommand{\theequation}{43\alph{equation}}
\begin{align}
\mathcal{P}_{3}: &\min_{\boldsymbol{p}_{t},\boldsymbol{f}_{t},\boldsymbol{\gamma}_t}\sum_{n=1}^N Q_{n,t} E_{n,t} \tag{35}\\
\text{s. t. } &\hyperref[p1]{(20b),(20e)-(20h)} \tag{35a}
\end{align}
\end{subequations}
In this subsection, we focus on the subproblem of controlling computational frequency $\boldsymbol{f}_t$ and transmission power $\boldsymbol{p}_{t}$ for UAVs given the UAV-EV association $\boldsymbol{\beta}_{t}$ and the allocated bandwidth $\boldsymbol{\gamma}_{t}$. Our goal is to optimize $\boldsymbol{f}_{t}$ and $\boldsymbol{p}_{t}$ of each UAV to minimize the objective function in \hyperref[eq35]{(35)}. At this point, the subproblem can be formulated as follows
\begin{subequations}
\renewcommand{\theequation}{35\alph{equation}}
\begin{align}
\bar{\mathcal{P}}_{3}: &\min_{\boldsymbol{p}_{t},\boldsymbol{f}_{t}}\sum_{n=1}^N Q_{n,t} E_{n,t} \tag{36}\\
\text{s. t. } &\hyperref[p1]{(20b),(20f),(20g)}\tag{36a}
\end{align}
\end{subequations}

Due to the coupling between the variables $\boldsymbol{f}_{t}$ and $\boldsymbol{p}_{t}$, the problem cannot be directly solved at this stage. We first solve for the optimal $\boldsymbol{f}_{t}$ given a fixed $\boldsymbol{p}_{t}$, and express $\boldsymbol{f}_{t}$ as a function of $\boldsymbol{p}_{t}$. This transforms the problem into a single-variable optimization problem, which can then be solved.
\subsubsection{Optimize $\boldsymbol{f}$ given $\boldsymbol{p}$}

Given parameters $ \boldsymbol{\beta}_{t} $, $\boldsymbol{\gamma}_t $, and $\boldsymbol{p}_t$, the transmission time $T_{n,t}^{\text{comm}}$ for each UAV is determined.
From formula \hyperref[eq10]{(10)}, we have
\begin{equation}
\frac{\partial E_{n,t}^{\text{comp}}}{\partial f_{n,t}}=2\varsigma K D_n C_{n,t}f_{n,t}> 0 
\end{equation}

To minimize energy consumption, each UAV should operate at the lowest feasible frequency. Consequently, the computational frequency for each UAV is set to satisfy 
\begin{equation}
  f_{n,t} = \frac{K L_b C_{n,t}}{T_{\text{max}} - T_{n,t}^{\text{comm}}}
\end{equation}
\subsubsection{Optimize $\boldsymbol{p}_t$}
By rearranging equation \hyperref[eq11]{(11)} we get $P_{n, t}=\frac{\gamma_{n, t} B N}{h_{n, t}}\left(2^{\frac{Z_{n,t}}{\gamma_{n,t} B T_{n, t}^{comm}}}-1\right)$, where $Z_{n,t}=\sum\nolimits_{m=1}^M\beta_{m,n}^tZ_m$.
Therefore, the training energy consumption and transmission energy consumption can be expressed as 
\begin{equation}
E_{n,t}^{comp}=\frac{\varsigma K^3 D_n^3 C_{n,t}^3}{(T-T_{n,t}^{comm})^2}
\end{equation}
and
\begin{equation}
E_{n,t}^{comm}=\frac{\gamma_{n, t} B NT_{n,t}^{comm}}{h_{n, t}}\left(2^{\frac{Z_{n,t}}{\gamma_{n,t} B T_{n, t}^{comm}}}-1\right)
\end{equation}
To ensure the UAV has sufficient time to complete model training, it must satisfy \( \frac{K L_b C_{m,n}}{f_{n}^{\text{max}}} \leq T - T_{n,t}^{\text{comm}},\forall n \). Therefore, the transmission power 
$p_n$ must satisfy
\begin{equation}
p_{n,t}\geq p_{n,t}^{min}=\frac{\gamma_{n, t} B N}{h_{n, t}}\left(2^{\frac{Z_{n,t}}{\gamma_{n,t} B \left (T-\frac{KD_nC_{n,t}} {f_{n}^{max}}\right )}}-1\right)
\end{equation}
Then, problem $\bar{\mathcal{P}}_{3}$ is transformed into
\begin{subequations}
\begin{align}
\hat{\mathcal{P}}_{3}: & \min_{\boldsymbol{p}_{t}}E_t= \sum_{n=1}^{N}Q_{n,t}E_{n,t}\tag{42}\\
\text{s. t. } &p_{n,t}^{min} \le p_{n,t}\leq p_{n,max}
\end{align}
\end{subequations}

It is easy to see that $E_{n,t}$ is a function of $T_{n,t}^{comm}$, and $T_{n,t}^{comm}$ is a function of $p_{n,t}$. So we have $\frac{\partial E_{n,t}}{\partial p_{n,t}}=\frac{\partial E_{n,t}}{\partial T_{n,t}^{comm}}\frac{\partial T_{n,t}^{comm}}{\partial p_{n,t}}$. We first study the properties of $E_{n,t}$ with respect to $T_{n,t}^{comm}$. The first-order
derivative of the objective function with respect to $T_{n,t}^{comm}$ is
\begin{equation}
\begin{aligned}
&\frac{\partial E_{n,t}}{\partial T_{n,t}^{comm}}  = \frac{2 \varsigma  K^3 D_n^3 C_{n,t}^3}{\left(T_{max}-T_{n,t}^{comm}\right)^3}\\
&\hspace{7mm}  + \frac{\gamma_{n,t} B N_0}{h_{n,t}}  \left( \left( 1-\frac{Z_{n,t} \ln 2}{\gamma_{n,t} B T_{n,t}^{comm}} \right) 2^{\frac{Z_{n,t}}{\gamma_{n,t} B T_{n,t}^{comm}}} - 1 \right)
\end{aligned}
\end{equation}
and the second-order derivative is
\begin{equation}
\frac{\partial^2 E_{n,t}}{\partial^2T_{n,t}^{comm}}=6\frac{\varsigma K^3D_n^3C_{n,t}^3}{(T_{n,t}^{comm})^4}+\frac{N_0Z_{n,t}^2(\ln2)^22^{\frac {Z_{n,t}}{\gamma_{n,t}BT_{n,t}^{comm}}}}{\gamma_{n,t}B(T_{n,t}^{comm})^3h_{n,t}}.
\end{equation}
Therefore, we have $\frac{\partial^2 E_{n,t}}{\partial^2 T_{n,t}^{\text{comm}}} > 0$ and $\frac{\partial^2 E_{n,t}}{\partial T_{n,t}^{\text{comm}} T_{m,t}^{\text{comm}}} = 0$ for $n \neq m$. It is straightforward to see that $\lim_{T_{n,t}^{\text{comm}} \to 0} \frac{\partial E_{t}}{\partial T_{n,t}^{\text{comm}}} = -\infty$ and $\lim_{T_{n,t}^{\text{comm}} \to T_{\text{max}}} \frac{\partial E_{n,t}}{\partial T_{n,t}^{\text{comm}}} = \infty$. Thus, there exists a unique $T_{n,t}^{\text{comm,*}}$ that minimizes $E_{n,t}$, i.e. $\frac{\partial E_{n,t}}{\partial T_{n,t}^{\text{comm,*}}} = 0$. Since $\frac{\partial T_{n,t}^{\text{comm}}}{\partial p_{n,t}} > 0$, this conclusion also applies to $p_{n,t}$, and the optimal value is denoted as $\hat{p}_{n,t}$. Taking into account restriction \hyperref[20h]{(20h)}, we obtain the optimal frequency and transmission power settings for each UAV in the following theorem.
\begin{theorem}\label{theorem2}
Given $\boldsymbol{\beta}_t$ and $\boldsymbol{\gamma}_t$, the optimal transmission power and frequency for each UAV are
\begin{equation}
\begin{aligned}
p_{n,t}^{*} = \begin{cases}
p_{n,t}^{\min}, & \hat{p}_{n,t} \leq p_{n,t}^{\min}, \\
p_{n,\max}, & \hat{p}_{n,t} \geq p_{n,\max}, \\
\hat{p}_{n,t}, & \text{otherwise}.
\end{cases}
\end{aligned}
\end{equation}
and
\begin{equation}
f_{n,t}^*=\frac{K L_b C_{n,t}}{T_{\text{max}} - T_{n,t}^{\text{comm,*}}}
\end{equation}
where $T_{n,t}^{\text{comm,*}}=\frac{Q_{n,t}}{\gamma_{n, t} B \log _2\left(1+\frac{\hat{p}_{n, t} h_{n, t}}{\gamma_{n,t}B N_0}\right)} $ and $\hat{p}_{n,t}$ corresponds to the solution of equation $\frac{\partial E_{n,t}}{\partial p_{n,t}}=0$.
\end{theorem}

\subsection{Optimal Bandwidth Allocation}
In this section, we will investigate how to determine the optimal bandwidth allocation \( \boldsymbol{\gamma}_t \).

Given \( \boldsymbol{\beta}_t \), \( \boldsymbol{f}_t \) and \( \boldsymbol{p}_t \), the training energy consumption $E_{n,t}^{comm}$ and training time $T_{n,t}^{comm}$ for each UAV are determined, which simplifies the objective function in problem $\mathcal{P}_{3}$ to be solely dependent on the transmission energy. Consequently, problem $\mathcal{P}_3$ can be reformulated as
\begin{subequations}\label{eq47}
\begin{align}
\widetilde{\mathcal{P}}_{3}:  &\min_{\boldsymbol{\gamma}_{t}}  E_{t}^{comm}=\sum_{n=1}^{N}\frac{Q_{n,t}p_{n,t}Z_{n,t}}{\gamma_{n,t}Blog_2\left (1+\frac{p_{n,t}h_{n,t}}{\gamma_{n,t}BN_0} \right )} \tag{47}\label{eq47}\\
\text{s. t. } & \gamma_{n,t} >  0\label{eq47a}\\
& \sum_{n=1}^{N}\gamma_{n,t} =  1\\
&\gamma_{n,t}Blog_2\left (1+\frac{p_{n,t}h_{n,t}}{\gamma_{n,t}BN_0} \right )\geq \frac{Z_{n,t}}{T-T_{n,t}^{comp}} 
\end{align}
\end{subequations}
From problem $\widetilde{\mathcal{P}}_{3}$, it can be seen that by appropriately allocating the bandwidth resources $\boldsymbol{\gamma}_t$, we can maximize the denominator in equation \hyperref[eq47]{(47)}, which is equivalent to minimizing its negative value, thus minimizing the objective function in $\widetilde{\mathcal{P}}_{3}$. Consequently, problem $\widetilde{\mathcal{P}}_{3}$ can be transformed into
\begin{subequations}
\begin{align}
\widetilde{\mathcal{P}}_{3}^{'}:  \min_{\boldsymbol{\gamma}_{t}} & \hat{E}_{t}^{comm} = -\sum_{n=1}^{N}\gamma_{n,t}Blog_2\left (1+\frac{p_{n,t}h_{n,t}}{\gamma_{n,t}BN_0} \right ) \tag{48}\\
\text{s. t. } &\hyperref[eq47]{(47a)-(47c)}
\end{align}
\end{subequations}
It is easy to see $\frac{\partial \hat{E}_{t}^{comm}}{\partial \gamma _{n,t}} = -Blog_2\left ( 1+\frac{p_{n,t}h_{n,t}}{\gamma_{n,t}BN_0}  \right )-\frac{p_{n,t}h_{n,t}B}{\left ( \gamma _{n,t}BN_0+p_{n,t}h_{n,t} \right )ln2} $ and $\frac{\partial^2 \hat{E}_{t}^{comm}}{\partial^2 \gamma _{n,t}} = \frac{p_{n,t}h_{n,t}B}{\left ( \gamma_{n,t}BN_0+p_{n,t}h_{n,t}\right )\gamma _{n,t}ln2}+\frac{p_{n,t}h_{n,t}B^2N_0}{\left ( \gamma_{n,t}BN_0+p_{n,t}h_{n,t}\right )^2ln2}>0, \forall n\in \mathcal{N}$. In addition, we have $\frac{\partial^2 \hat{E}_{t}^{comm}}{\partial \gamma _{n,t}\partial\gamma _{j,t}}$ for $n\neq j$. Thus, the Hessian matrix of $\hat{E}_{t}^{comm}$ with respect to $\gamma_{n,t}$ is a diagonal matrix with positive entries on the diagonal, indicating that it is positive definite. Therefore, the objective function is convex with respect to $\gamma_{n,t}$. Given that all constraint conditions are also convex with respect to $\gamma_{n,t}$, problem $\widetilde{\mathcal{P}}_{3}^{'}$ is therefore a convex optimization problem. Next, we use the method of Lagrange multipliers to transform the constrained optimization problem $\widetilde{\mathcal{P}}_{3}^{'}$ into an unconstrained optimization problem.

The Lagrangian function of problem $\widetilde{\mathcal{P}}_{3}^{'}$ is defined as
\begin{equation}
\begin{aligned}&\mathcal{L}(\boldsymbol{\gamma}_{t},\lambda, \mu, \varrho )=\sum_{n=1}^{N}\left(-\gamma_{n,t}B\log_{2}\left(1+\frac{p_{n,t}h_{n,t}}{\gamma_{n,t}BN_{0}}\right)\right)\\
&\hspace{1cm}+\lambda\left(\frac{Z_{n,t}}{T-T_{n,t}^{comp}}-\gamma_{n,t}B\log_{2}\left(1+\frac{p_{n,t}h_{n,t}}{\gamma_{n,t}BN_{0}}\right)\right)\\
&\hspace{1cm}+\mu\left(\sum_{n=1}^N\gamma_{n,t}-B\right)+ \varrho \gamma_{n,t}\end{aligned}
\end{equation}
where $\lambda, \mu$ and $\varrho$ are Lagrange multipliers. By utilizing the Karush-Kuhn-Tucker (KKT) condition, we have
\begin{subequations}\label{eq:problem}
\begin{align}
& \sum_{n=1}^{N} \gamma_{n,t} = B, \label{50a} \\
& \frac{Z_{n,t}}{T-T_{n,t}^{comp}}-\gamma_{n,t}B\log_{2}\left(1+\frac{p_{n,t}h_{n,t}}{\gamma_{n,t}BN_{0}}\right) \leq 0, \label{50b} \\
& \gamma_{n,t} > 0, \ \lambda \geq 0, \ \varrho \geq 0, \label{50c} \\
& \lambda\left(\frac{Z_{n,t}}{T-T_{n,t}^{comp}}-\gamma_{n,t}B\log_{2}\left(1+\frac{p_{n,t}h_{n,t}}{\gamma_{n,t}BN_{0}}\right)\right) = 0, \label{50d} \\
& \varrho \gamma_{n,t}= 0, \label{50e} \\
& \mu + \varrho + (1 + \lambda) \times \left[ \frac{p_{n,t}h_{n,t}}{\gamma_{n,t}BN_0} \frac{1}{\left(1 + \frac{p_{n,t}h_{n,t}}{\gamma_{n,t}B N_0}\right)\ln 2} \right. \notag\\
& \quad \left.- \log_2\left(1 + \frac{p_{n,t}h_{n,t}}{\gamma_{n,t}B N_0}\right)\right] = 0. \label{50f}
\end{align}
\end{subequations}
We solve the equations to obtain the optimal bandwidth allocation for UAV, which can be divided into two cases:

$\bullet \, \lambda > 0$: From \hyperlink{50d}{(50d)}, we can see that
\begin{equation}
\left( \frac{Z_{n,t}}{T - T_{n,t}^{comp}} - \gamma_{n,t}B \log_2 \left( 1 + \frac{p_{n,t} h_{n,t}}{\gamma_{n,t} B N_0} \right) \right) = 0,\forall n 
\end{equation}
Through similar derivation as in \cite{b49}, we have
\begin{equation}
    \gamma _{n,t}=\frac{-Q_{n,t} ln 2}{\left (\Delta T-T_{n,t}^{comp}\right )\left [\mathcal{W}\left (-\chi_{n,t} e^{-\chi_{n,t}  } \right ) +\chi_{n,t}\right ]}
\end{equation}
where  $\mathcal{W(\cdot)}$ is the Lambert-W function and $\chi_{n,t}=\frac{d_{n,t}N_0}{\left(\Delta T-T_{n,t}^{comp}\right )P_{n,t}h_{n,t}^2}$.

$\bullet \, \lambda = 0:$ In this case, we turn to equation \hyperlink{50f}{(50f)}. From equation \hyperlink{50e}{(50e)}, since $\gamma_{n,t} > 0$ always holds, we have $\varrho = 0$. Therefore, equation \hyperlink{50f}{(50f)} can be rewritten as
\begin{equation}\label{eq53}
\mu+\frac{p_{n,t}h_{n,t}}{\gamma_{n,t}BN_0} \frac{1}{\left(1 + \frac{p_{n,t}h_{n,t}}{\gamma_{n,t}B N_0}\right)\ln 2}- \log_2\left(1 + \frac{p_{n,t}h_{n,t}}{\gamma_{n,t}B N_0}\right)=0
\end{equation}
Denoting $A=\frac{p_{n,t}h_{n,t}}{\gamma_{n,t}BN_0}$, then equation \hyperref[eq53]{(53)} can be rewritten as $\mu+\frac{A}{(1+A)ln2}-log_2(1+A)=0 $. By using the logarithm change of base formula and transforming the equation, we have $ln(1+A)=\mu ln2+\frac{A}{1+A}$. Taking the exponential of both sides with the natural base, the result is $1+A=e^{ln2(\mu+\frac{1}{ln2})-\frac{1}{1+A} }$. Then, we have $-2^{-(\mu+\frac{1}{ln2} )}=-\frac{1}{1+A}e^{-\frac{1}{1+A} } $. By using the Lambert-W function, we obtain the optimal bandwidth allocation for each UAV $n$:
\begin{equation}
\gamma_{n,t}^*\left ({\mu}\right )=-\frac{p_{n,t}h_{n,t}}{BN_0}\frac{\mathcal{W}\left(-2^{-{(\mu+\frac{1}{\ln2})}}\right)}{1+\mathcal{W}\left(-2^{-{(\mu+\frac{1}{\ln2})}}\right)}.
\end{equation}
It should be noted that the optimal Lagrange multiplier $\mu$ is solved by bisection to satisfy constraint \hyperref[eq46]{(46b)}. The details are omitted here for simplicity.
\subsection{Optimal UAV-EV Association Algorithm}
In the previous section, we obtained the methods to get the optimal $\boldsymbol{f}_t$, $\boldsymbol{p}_t$, and $\boldsymbol{\gamma}_t$ given $\boldsymbol{\beta}_t$. Next, we first solve problem \( \mathcal{P}_{3} \) by jointly optimizing $\boldsymbol{f}_t$, $\boldsymbol{p}_t$, and $\boldsymbol{\gamma}_t$, and then design an efficient UAV-EV association strategy.

Based on the previous analysis, both problem $\hat{\mathcal{P}}_{3}$ and problem $\widetilde{\mathcal{P}}_{3}$ are convex problems. Therefore, the joint optimization problem of $\boldsymbol{f}_t$, $\boldsymbol{p}_t$, and $\boldsymbol{\gamma}_t$, i.e., problem $\mathcal{P}_{3}$, is also a convex problem\cite{b41}. Therefore, we can solve it iteratively using the block coordinate descent (BCD) method. Each iteration of the block coordinate descent method consists of two steps: solving the bandwidth allocation $\boldsymbol{\gamma}_t$ given $\boldsymbol{f}_t$ and $\boldsymbol{p}_t$, and solving $\boldsymbol{f}_t$ and $\boldsymbol{p}_t$ given $\boldsymbol{\gamma}_t$. The iterations proceed until convergence. The block coordinate descent method's iteration steps are summarized in Algorithm \hyperref[algorithm1]{1}.
\begin{algorithm}\label{algorithm1}
\caption{Joint Optimization of $\boldsymbol{f}$, $\boldsymbol{p}$, and $\boldsymbol{\gamma}$ through Block Coordinate Descent}
\begin{algorithmic}[1]
\State \textbf{Input:} Initial feasible calculation frequency $\boldsymbol{f}_0$, transmission power $\boldsymbol{p}_0$, and bandwidth allocation ratio $\boldsymbol{\gamma}_0$, the tolerance $\epsilon$
\State \textbf{Output:} The optimized $\boldsymbol{f}$, $\boldsymbol{p}$, and $\boldsymbol{\gamma}$
\State Set iteration = 0
\Repeat
    \State $i \gets i + 1$
    \State \textbf{Given} $\boldsymbol{\gamma_t}$, solve problem $\hat{\mathcal{P}}_{3}$ to obtain $\hat{\boldsymbol{p}}_{t}$:
    \[
    \boldsymbol{p}_t = \arg \min U(\boldsymbol{f}, \boldsymbol{p}, \boldsymbol{\gamma_t})
    \]
    \State calculate
    \[
    p_{n,t} = \begin{cases}
        p_{n,t}^{\min}, & \hat{p}_{n,t} \leq p_{n,t}^{\min}, \\
        p_{n,t}^{\max}, & \hat{p}_{n,t} \geq p_{n,t}^{\max}, \\
        \hat{p}_{n,t}, & \text{otherwise}.
    \end{cases}
    \]
    and \hspace{0.5cm} $f_{n,t} = \frac{K L_b C_{n,t}}{T_{\text{max}} - T_{n,t}^{\text{comm,*}}}$
    \State \textbf{Given} $\boldsymbol{f_t}$ and $\boldsymbol{p_t}$, solve problem $\widetilde{\mathcal{P}}_{3}$ to obtain $\boldsymbol{\gamma}_{t}$:
    \[
    \boldsymbol{\gamma}_t = \arg \min \mathcal{L}(\boldsymbol{\gamma}_{t},\lambda, \mu, \varrho )
    \]
    \State Calculate $\Delta=U_i-U_{i-1}$
\Until{$\Delta \leq \epsilon$}
\State \textbf{Return:} Optimized $\boldsymbol{f}_t$, $\boldsymbol{p}_t$, and $\boldsymbol{\gamma}_t$ 
\end{algorithmic}
\end{algorithm}
Through Algorithm \hyperref[algorithm1]{1}, we can obtain the optimal settings for $\boldsymbol{f}_t$, $\boldsymbol{p}_t$, and $\boldsymbol{\gamma}_t$ under a given UAV-EV association $\boldsymbol{\beta}_t$. Based on this, we will now determine the optimal UAV-EV association. 
The authors in \cite{b44} and \cite{b50} utilized the Age of Updates (AoU) and channel gain as key metrics for edge server-client association, respectively. However, both studies failed to account for the UAV's total energy constraint. To address this limitation, we draw inspiration from the SEA algorithm \cite{b39} and propose an efficient UAV-EV association algorithm described as follows.

According to \hyperref[eq21]{(21)}, the UAV set can be classified into two categories based on whether \(Q_{n,t} > 0\) or \(Q_{n,t} = 0\). We perform the optimal association for these two categories separately. From equation \hyperref[eq27]{(27)}, we notice that UAVs with \(Q_{n,t} = 0\) do not affect the first term in \hyperref[eq27]{(27)}. Therefore, to minimize \hyperref[eq27]{(27)}, UAVs with \(Q_{n,t} = 0\) should be matched with the tasks that have the largest \(\alpha_m^t\), which means that UAVs with relatively abundant energy should support the most important tasks currently. Next, we determine the associations for UAVs with \(Q_{n,t} > 0\). According to equation \hyperref[eq27]{(27)}, each UAV is inclined to associate with the EV that yields a lower \( E_{n,t} \) and a higher \( \alpha_m^t \). We first identify the association preferences of each UAV and then introduce a two-stage algorithm designed to optimize the UAV association for those with \( Q_{n,t} > 0 \), adhering to the constraint \hyperref[20e]{(20e)}. Specifically, we first define the utility function for each UAV as
\begin{equation}
J_n(\boldsymbol{\beta}_{n},p_{n},f_{n},\gamma_{n})=Q_{n,t} E_{n,t} - V \sum_{m = 1}^{M} \alpha_m^t \sum_{n = 1}^{N} \beta_{m,n}^t D_n 
\end{equation}
In the first stage, each UAV $n$ is allocated the same bandwidth ratio $\gamma_{n,t} = \frac{1}{N},\forall n$. Subsequently, each UAV is individually associated with each EV $m$ and solves problem $\bar{\mathcal{P}}_{3}$, with the corresponding utility function obtained as $J_{n,m}=J_n(\beta_{m,n}=1,p_{n}^*,f_{n}^*,1/N )$. After that, each UAV with $Q_{n,t} > 0$ is associated with the EV $m = \arg\min_{m} U_{n,m}$. In the second stage, adjustments are made to ensure that $\sum_{n=1}^N\beta_{m,n}^t\geq \delta,\forall m$. To facilitate UAV transfers between EVs while minimizing the change in the utility function before and after the transfer, we calculate the transfer utility loss for each UAV in each EV with $\sum_{n=1}^N\beta_{m,n}^t> \delta$ as \( \Delta_{n}^{i,j} = J_{n,j} - J_{n,i} \), which represents the increase in the objective function when UAV \( n \) is transferred from EV \( i \) to EV \( j \). Furthermore, each EV with $\sum_{n=1}^N\beta_{m,n}^t> \delta$ can transfer a maximum of $ \hbar_n = \sum_{n=1}^N\beta_{m,n}^t- \delta$ UAVs. Each EV \( n \) adds the smallest \( \hbar_n \) values of \( \Delta_{n}^{i,j} \) to \( \Lambda \). Then, \( \Lambda \) is sorted in ascending order, and the transfer process begins based on \( \Delta_{n}^{i,j} \) from smallest to largest.
When UAV \( n \) is transferred, all the \( \Delta_{n}^{i,j} \) values related to \( n \) are removed from \( \Lambda \). The transfer process is then repeated until each EV with \( \sum_{n=1}^N \beta_{m,n}^t < \delta \) satisfies \( \sum_{n=1}^N \beta_{m,n}^t = \delta \).

\subsection{Performance and Complexity Analysis}
Previously, we developed the optimal UAV-EV association and resource allocation strategy. In this section, we will analyze the performance and complexity of the proposed algorithm.
\begin{theorem}\label{theorem3}
The proposed algorithm satisfies the following properties

\noindent(a)The performance gap between our proposed strategy and the optimal offline strategy satisfies
\begin{equation}
\sum_{t=0}^{T} U^{t,*} - \sum_{t=0}^{T} \hat{U}^t \leq \frac{T^2B}{V} 
\end{equation}
where \( U^t \) is defined in \hyperref[eq25]{(25)} and \( U^{t,*} \) represents the optimal utility achieved under the optimal offline strategy.

\noindent(b)The difference between the energy consumed by our algorithm and the energy budget satisfies
\begin{equation}
\frac{1}{T}\sum_{t=0}^{T-1}\sum_{n=1}^{N}E_{n,t} \leq \sum_{n=1}^{N}\bar{E}_n+   \sqrt{\frac{2B}{T}+\frac{2V\sum_{t=0}^{T-1}U^{t,*}}{T^2}} 
\end{equation}
\end{theorem}
\noindent\textit{proof}: See Appendix \hyperref[AppendixC]{C} in the supplementary material.

Theorem \hyperref[theorem3]{3} presents the performance upper bounds of the proposed UAV-EV association and resource allocation strategy, demonstrating that the algorithm satisfies the total energy constraint of each UAV within a bounded factor of \( O(\sqrt{V}) \). Additionally, the algorithm achieves \( O(1/V) \) optimality in enhancing multi-task performance. In other words, there exists a trade-off between energy consumption and multi-task performance, characterized by the \( O(\sqrt{V}, 1/V) \) relationship. Specifically, as \( V \) increases, greater emphasis is placed on multi-task performance, while a smaller \( V \) prioritizes the UAV's energy consumption limit.

The computational complexity of calculating the Shapley value for tasks is \( O(M \cdot 2^M )\)\cite{b42}, which is acceptable given that \( M \) is relatively small. For bandwidth allocation, the bisection method has a computational complexity of \( O(\log_2(\frac{ub-lb}{\varepsilon  } )) \).  Since UAV-EV association requires each UAV to execute \( M \) instances of BCD, the total computational complexity is \( MN\cdot\mathcal{O}(BCD)  \cdot \mathcal{O}(\log_2(\frac{ub-lb}{\varepsilon  } ))  + \mathcal{O}(M \cdot 2^M) \). Where $ub$ and $lb$ represent the initial boundary for the binary search method. The computational complexity of the exhaustive search method is \(M^N\cdot\mathcal{O}(BCD)  \cdot \mathcal{O}(\log_2(\frac{ub-lb}{\varepsilon  } ))  + \mathcal{O}(M \cdot 2^M)\), which is significantly higher than that of the proposed algorithm.


\section{Numerical Results}
\subsection{Datasets and local models}
This study leverages the MNIST, FMNIST, and UAV image dataset FLAME\cite{b46}, modifying them to suit specific scenarios considered in the research. Four tasks are constructed: (1) \textbf{RC-MNIST} assigns distinct colors to MNIST digits and their backgrounds, with colors selected from a predefined set of 10 RGB values, and rotates the digits at angles from the vector \([0, 36, \ldots, 324]\). This task supports digit color recognition, rotation angle recognition, and digit classification. (2) \textbf{SD-MNIST} is derived from the MNIST dataset by extracting digits 0 to 5, arranging two images side by side, and assigning different colors to each digit and its background. It is designed for four tasks: addition and subtraction of the two digits, as well as identifying the colors of the left and right digits. (3) \textbf{RC-FMNIST} applies similar transformations as RC-MNIST to fashion items and backgrounds in FMNIST dataset, with item rotation included. (4) \textbf{FLAME} dataset contains real images captured from UAV perspectives at different heights and angles. It is used to accomplish situational awareness tasks, including scene recognition and fire detection. For these tasks, Dirichlet parameters \(\alpha_1\) and \(\alpha_2\) are utilized to control the UAV sample quantity distribution and class distribution, respectively.

For the RC-MNIST, SD-MNIST, and RC-FMNIST datasets, the color recognition task uses a two-layer CNN with 167,690 parameters and requires 1,481,792 FLOPs per sample, while other tasks use a three-layer CNN with 249,738 parameters and 1,563,648 FLOPs per sample. Both configurations include three fully connected layers, with the gradients of the first two convolutional layers shared. For the FLAME dataset, both tasks are handled by a four-layer CNN, where the first two convolutional layers can share gradients, followed by two fully connected layers. This model has 521,443 parameters and requires 211,237,152 FLOPs. Each CPU cycle can process 8 FLOPs.
\subsection{Hyperparameters and baselines}
The remaining parameters in the wireless network are shown in Table \hyperref[table2]{II}. To evaluate the performance of the proposed algorithm, we compare it with three baseline strategies:

\textbf{AoU-Based}\cite{b44}: The Age of Update(AoU) is used to measure the importance of UAV $n$ to task $m$. In each round, the UAV associates with the EV that has the maximum corresponding AoU, without sharing task-related knowledge.

\textbf{Channel Aware}\cite{b50}: Each UAV associates with the EV having the optimal channel gain, thereby reducing transmission resource consumption, without task knowledge sharing.

\textbf{FedAvg}\cite{b45}: UAVs and EVs are randomly associated, and similarly, no knowledge sharing is performed.

In the following, we compare the proposed algorithm, which incorporates task knowledge sharing and optimal UAV-EV association, with three baseline methods. The algorithms are categorized into two groups: \textbf{Intimacy-Based} and \textbf{All-Shared}. The former performs knowledge sharing based on task affinity, while the latter uniformly shares knowledge across all tasks without considering task relevance.

\addtolength{\topmargin}{0.03in}
\begin{table}\label{table2}
\begin{center}
\caption{Parameter settings}
\begin{tabu} to 0.45\textwidth {X[c]|X[b]|X[c]|X[m]}
\hline
Parameter & Value & Parameter & Value \\
\hline
$N$ & 10 & $f_{max}$ & 2GHz \\
$B$ & 10MHz &$L_b$& 64 \\
$N_0$ & -174dBm/Hz & $H$ & [100,150]m \\
$\eta$ & 0.01 & $\kappa$,$\varpi$,$\ell$ &0.8\\
$\varpi$ & 0.8&  & \\
\hline
\end{tabu}
\end{center}
\end{table}

\subsection{Performance of Knowledge Sharing}
In this part, we compare the performance of the proposed scheme across four tasks with $N = 10$. For each task, the Dirichlet parameters are set as $\alpha_1=1$ and $\alpha_2=1$. It should be noted that this part focuses solely on task performance without considering resource constraints. 
\begin{figure}[htb]\label{fig3}
	\centering
	\begin{minipage}{0.49\linewidth}\label{fig5a}
		\centering
		\includegraphics[width=\textwidth]{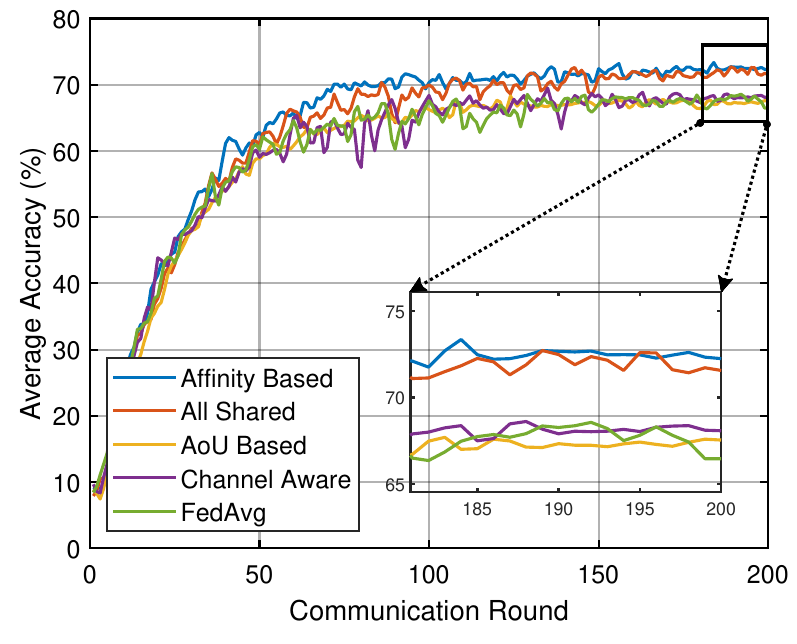}
		\subcaption{RC-MNIST: 2Task}
	\end{minipage}
	\begin{minipage}{0.49\linewidth}\label{fig5b}
		\centering
		\includegraphics[width=\textwidth]{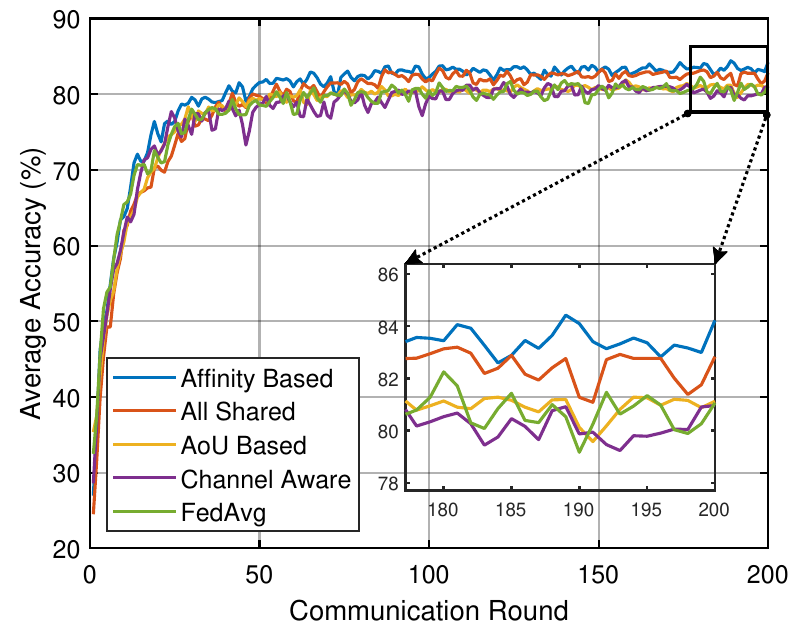}
		\subcaption{RC-MNIST: 3Task}
	\end{minipage}
	\caption{Performance Comparison on SD-MNIST and RC-FMNIST}
	\label{fig:avg_acc_variance}
\end{figure}
Fig. \hyperref[fig3]{3(a)} illustrates the results of rotation angle recognition and digit recognition tasks on the RC-MNIST. Due to the improved generalization performance brought by knowledge sharing, both the Affinity-Based and All-Shared strategies achieve faster convergence and higher final accuracy, outperforming the best baseline by 3.11\% and 3.77\%, respectively. 
Fig. \hyperref[fig3]{3(b)} presents the results of rotation angle recognition, digit recognition, and color recognition tasks using the RC-MNIST dataset. All Shared and Affinity Based achieve convergence accuracy improvements of 1.12\% and 2.55\% over the best baseline, respectively. While All Shared exhibits slower training in the early stages due to task conflicts, Affinity Based strategy effectively captures task correlations, achieving the fastest training speed and the highest convergence accuracy.

\begin{figure}[htb]\label{fig4}
	\centering
	\begin{minipage}{0.49\linewidth}\label{fig4a}
		\centering
		\includegraphics[width=\textwidth]{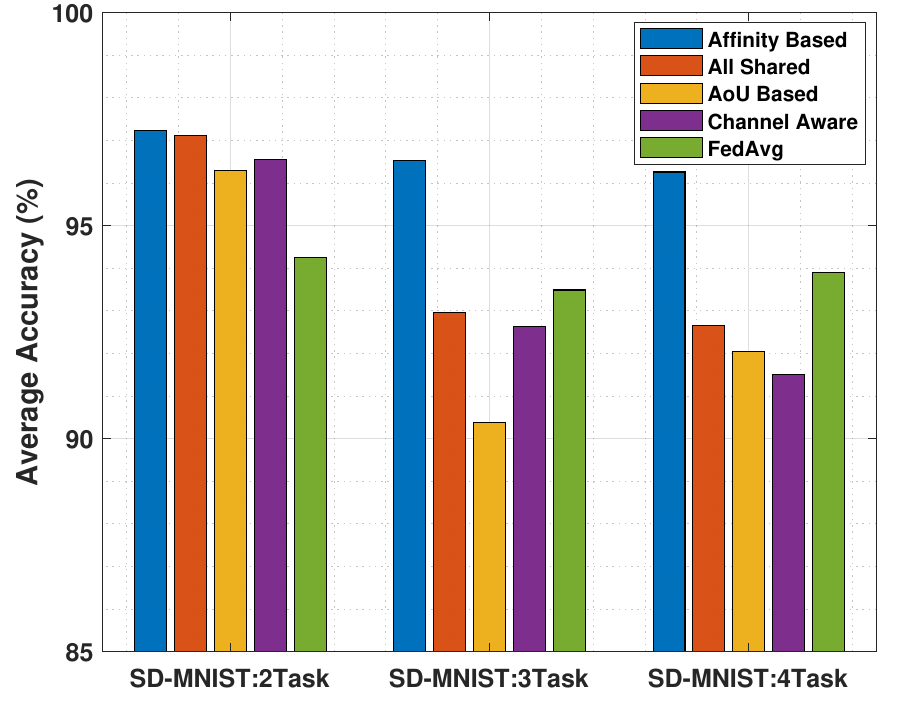}
		\subcaption{SD-MNIST}
	\end{minipage}
	\begin{minipage}{0.47\linewidth}\label{fig4b}
		\centering
		\includegraphics[width=\textwidth]{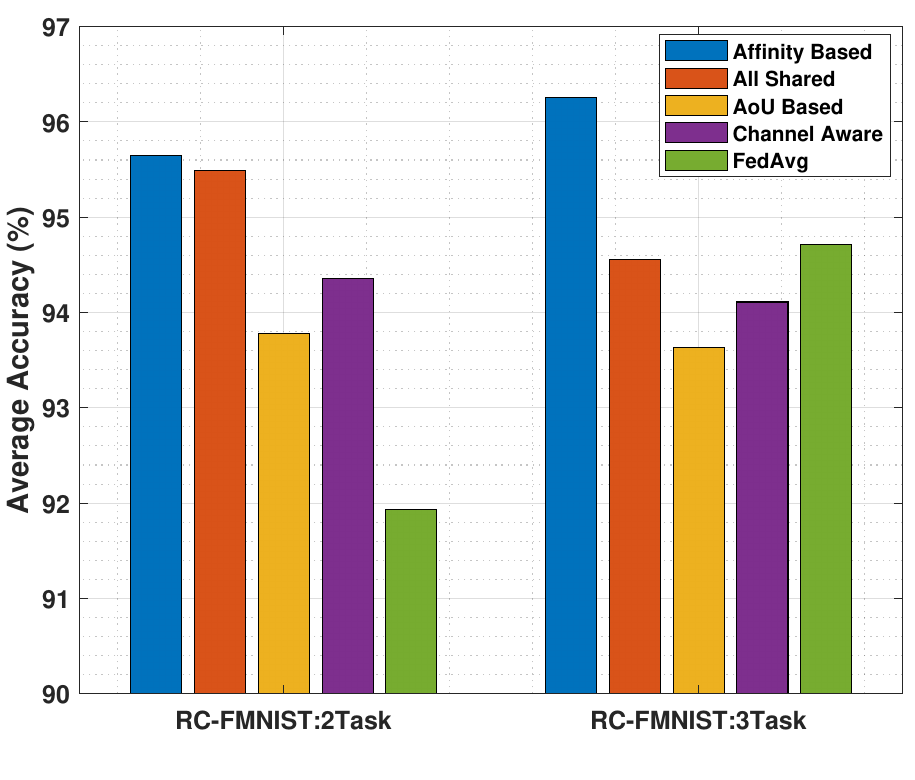}
		\subcaption{RD-FMNIST}
	\end{minipage}
 
	\caption{Performance Comparison on SD-MNIST and RC-FMNIST}
	\label{fig:avg_acc_variance}
\end{figure}
 
Fig. \hyperref[fig4a]{4(a)} illustrates the final convergence accuracy for three task groups performed on SD-MNIST. Specifically, SD-MNIST: 2Task involves two fundamental tasks: calculating the sum and difference of digits, while SD-MNIST: 3Task and SD-MNIST: 4Task incorporate the identification of the left and right digit colors, respectively. The Affinity-Based Strategy outperforms the highest baseline by 0.69\%, 3.04\%, and 2.36\% for these tasks. In contrast, due to task conflicts, the performance of the All-Shared Strategy may degrade, falling below certain baselines. Fig. \hyperref[fig4b]{4(b)} presents the results of performing two task groups on RC-FMNIST. Compared to the best baseline, the Affinity-Based method achieves final accuracy improvements of 1.29\% and 1.55\% in RC-FMNIST:2Task and RC-FMNIST:3Task respectively. Additionally, the All-Shared strategy demonstrates a performance gain of 1.13\% over the baseline for fully correlated tasks. However, it underperforms compared to FedAvg when handling conflicting tasks.
    
    

\begin{figure}[h]\label{fig5}
    \centering
    \includegraphics[width=0.49\textwidth]{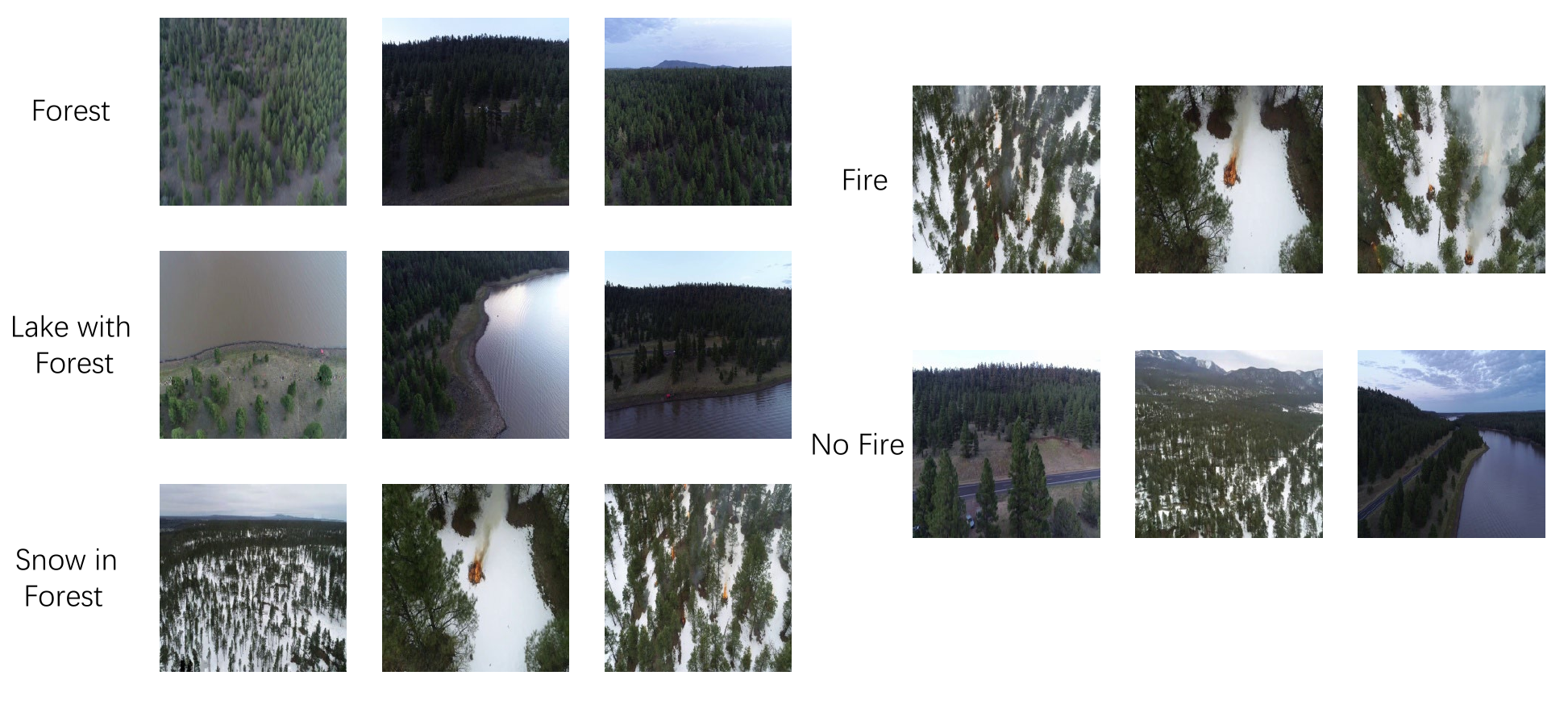}  
    \caption{Reconstructed FLAME Dataset}
\end{figure}

\begin{figure}[h]\label{fig6}
    \centering
    \includegraphics[width=0.45\textwidth]{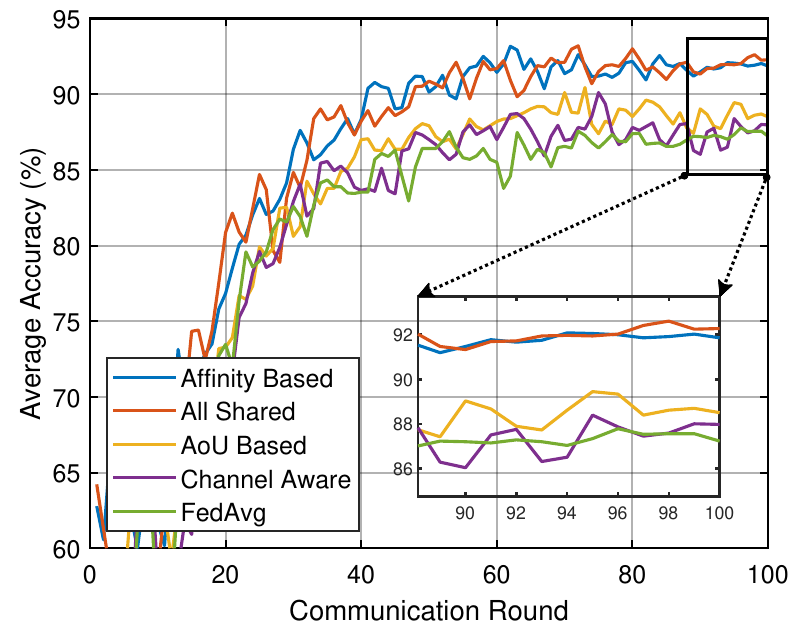}
    \caption{Performance Comparison on FLAME Dataset}
\end{figure}

In the following, we use the FLAME dataset, which contains images captured by UAVs from various altitudes and viewpoints, as well as some ground-level perspectives. These ground-level images are used as EV data. Initially designed for wildfire detection, we have restructured the dataset to address two specific challenges: scene recognition and wildfire monitoring, as shown in Fig.\hyperref[fig5]{5}. Scene recognition is used to identify three different scenarios: forest, lake with forest, and snow in forest. Wildfire monitoring focuses on classifying whether a fire is present or not. 
The results in Fig. \hyperref[fig6]{6} demonstrate that knowledge sharing improves the generalization capability of the model. Specifically, the \textbf{Affinity-Based} and \textbf{All-Shared} strategies achieve accuracy improvements of 3.44\% and 3.7\%, respectively. This validates the effectiveness of related task knowledge sharing strategy on the FLAME dataset.

\subsection{Performance of Proposed Algorithm}
\begin{figure}[htb]\label{fig7}
	\centering
	\begin{minipage}{0.49\linewidth}\label{fig7a}
		\centering
		\includegraphics[width=\textwidth]{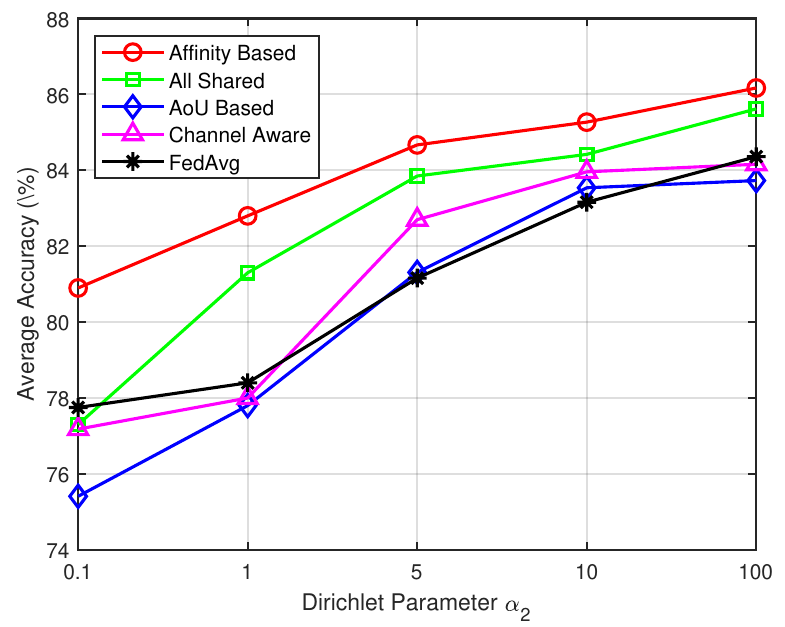}
		\subcaption{RC-MNIST: 3Task}
	\end{minipage}
	\begin{minipage}{0.49\linewidth}\label{fig7b}
		\centering
		\includegraphics[width=\textwidth]{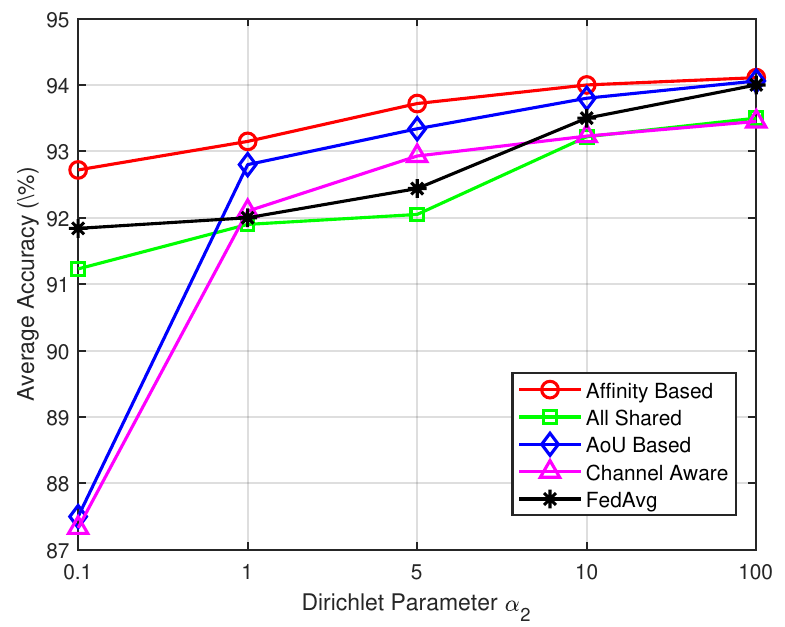}
		\subcaption{FLAME}
	\end{minipage}
	\caption{Performance Comparison under Different Non-IID Levels}
	\label{fig:avg_acc_variance}
\end{figure}

We compare the performance of the proposed algorithm under different levels of non-IID (niid) conditions on the RC-MNIST:3Task and FLAME dataset. We set $\alpha_1=1$ and vary $\alpha_2$ across the values $\{0.1, 1, 5, 10, 100\}$. Fig.\hyperref[fig7]{7(a)} and Fig.\hyperref[fig7]{7(b)} show the average accuracy of five different strategies after 200 training epochs on RC-MNIST: 3Task and 100 training epochs on FLAME, respectively, under varying values of \( \alpha_2 \). The results indicate that the Affinity-Based strategy achieves the highest accuracy across all five values of \( \alpha_2 \). Specifically, it improves the highest baseline accuracy by 3.1\%, 4.8\%, 1.98\%, 1.31\%, and 1.91\% on RC-MNIST: 3Task, and by 0.88\%, 0.35\%, 0.38\%, 0.2\%, and 0.05\% on FLAME. The All-Shared strategy, due to potential conflicts between tasks, performs worse than some baseline strategies in certain cases.

\begin{figure}[h]\label{fig8}
    \centering
    \includegraphics[width=0.45\textwidth]{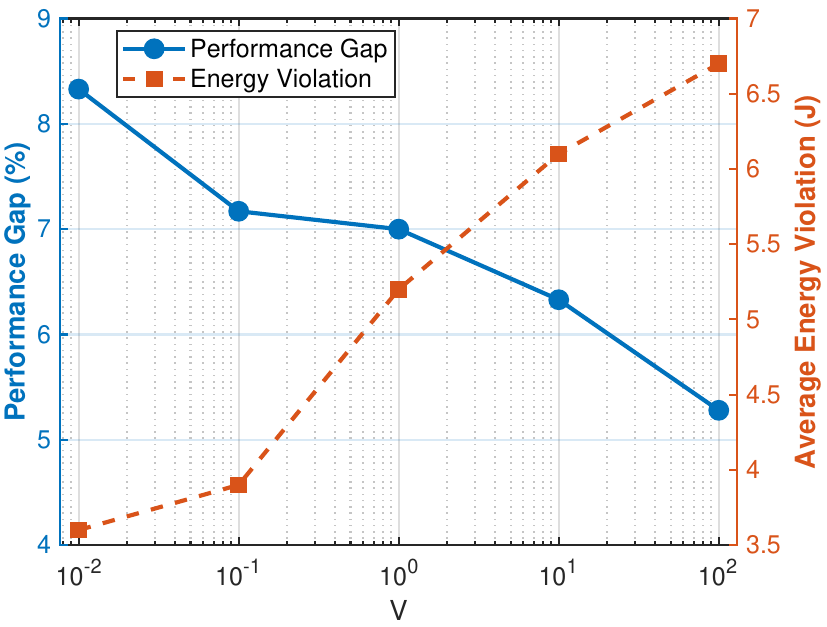}  
    \caption{Tradeoff Between Task Performance Gap and Energy Violation}
\end{figure}

Next, we investigate the impact of the parameter \( V \) on both the performance gap and the energy consumption of UAVs. Since the optimal task performance cannot be obtained, we define the performance gap as 100 minus the average task accuracy achieved. The experiments were conducted on the FLAME dataset over 100 iterations with \( \alpha_1 = 1 \) and \( \alpha_2 = 1 \). The results were tested for five different values of \( V \in \{0.01, 0.1, 1, 10, 100\} \). The results in Fig.\hyperref[fig8]{8} indicate that as \( V \) increases, the average task accuracy after 100 iterations also improves, i.e., the performance gap decreases, but the average energy violation per UAV also rises. This suggests that higher values of \( V \) tend to improve task performance at the cost of higher energy consumption, which is consistent with the theoretical analysis in Theorem \hyperref[theorem3]{3}.


\begin{figure}[htb]\label{fig9}
	\centering
	\begin{minipage}{0.49\linewidth}\label{fig9a}
		\centering
		\includegraphics[width=\textwidth]{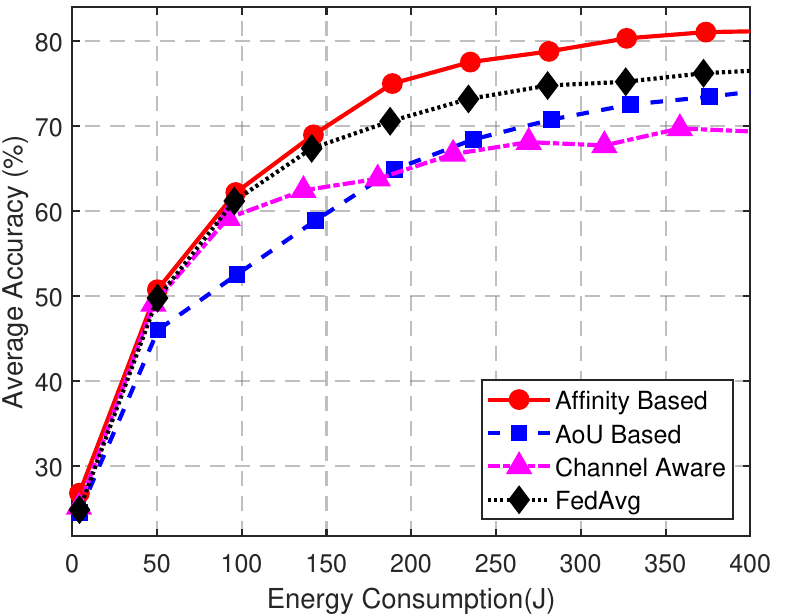}
		\subcaption{RC-MNIST: 3Task}
	\end{minipage}
	\begin{minipage}{0.49\linewidth}\label{fig9b}
		\centering
		\includegraphics[width=\textwidth]{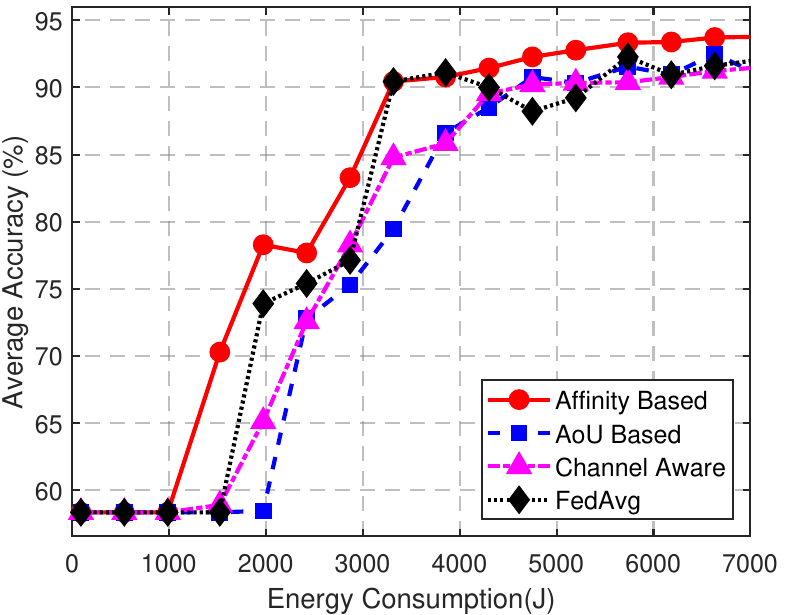}
		\subcaption{FLAME}
	\end{minipage}
	\caption{Average Accuracy versus Energy Consumption}
	\label{fig:avg_acc_variance}
\end{figure}

We further evaluate the resource efficiency of the proposed algorithm on the RC-MNIST: 3Task and FLAME datasets. For RC-MNIST: 3Task, the energy consumption limit per round is set to \( \bar{E}_{n,t} = 0.4 \, \text{J}, \, \forall n, t \), and for FLAME, the limit is 9 J, resulting in total energy budgets of \( E_{\text{total}} = 0.4T \, \text{J} \) and \( 9T \, \text{J} \) for each UAV, respectively. The time budget per round is constrained to \( T_{\text{max}} = 3 \, \text{s} \) for RC-MNIST: 3Task and \( T_{\text{max}} = 10 \, \text{s} \) for FLAME. As illustrated in Fig.\hyperref[fig9]{9}, we compare the performance of various algorithms as a function of energy consumption. It is important to note that all baseline strategies were optimized for resource allocation through Problem \( \mathcal{P}_3 \). The results presented in Fig.\hyperref[fig9]{9} indicate that the Affinity-Based scheme, with optimal UAV-EV association and resource allocation, achieves superior efficiency. Specifically, it outperforms the baseline strategies, yielding accuracy improvements of 6.87\%, 12.69\%, and 4.65\% for the RC-MNIST: 3Task dataset at an energy consumption of 400 J, and 1.66\%, 2.16\%, and 1.11\% for the FLAME dataset at 7000 J.

\section{Conclusion}
In this paper, we propose a novel UAV swarm-based multi-task federated learning scheme, leveraging data from various perspectives by UAVs to perform multiple tasks simultaneously. To address the challenges posed by heterogeneous and time-varying task demands, we introduce a task attention mechanism to measure the dynamic importance of tasks, along with a task affinity metric to capture task correlations. These mechanisms enable precise knowledge sharing across tasks, thereby improving convergence speed and generalization performance. To further enhance overall performance under resource constraints, we derive closed-form expressions for the optimal transmission power, computation frequency, and bandwidth allocation of UAVs. Building on this, we develop an efficient two-stage algorithm to determine the optimal UAV-EV association. Through theoretical analysis, we demonstrate that the proposed algorithm achieves an \( O(\sqrt{V},1/V) \) trade-off between task performance and energy consumption. Extensive simulation results further validate the effectiveness of the proposed scheme.

\section{Appendix}
\subsection{Proof of Lemma 1}\label{AppendixA}
We first prove the property of the function \( F_m(\boldsymbol{w}_m^s, \boldsymbol{w}_m^u) \). According to assumption 2, we have
\begin{equation}
\begin{aligned}
&||\nabla_s F_{m}(\bar{\boldsymbol{w}}^{s}, \boldsymbol{w}_{m}^{u})-\nabla_s F_{m}(\check{\boldsymbol{w}}^{s},\boldsymbol{w}_{m}^{u})||\\
&=|| \sum_{n=1}^{N} \frac{D_n}{D} \nabla_s F_{m,n}(\bar{\boldsymbol{w}}^{s}, \boldsymbol{w}_{m}^{u})\!-\!\sum_{n=1}^{N} \frac{D_n}{D} \nabla_s F_{m,n}(\check{\boldsymbol{w}}^{s},\boldsymbol{w}_{m}^{u})||\\
&\leq \sum_{n=1}^{N} \frac{D_n}{D}||\nabla_s F_{m,n}(\bar{\boldsymbol{w}}^{s}, \boldsymbol{w}_{m}^{u})-\nabla_s F_{m,n}(\check{\boldsymbol{w}}^{s},\boldsymbol{w}_{m}^{u})||\\
&\leq L_s||\bar{\boldsymbol{w}}^{s}-\check{\boldsymbol{w}}^{s}||
\end{aligned}
\end{equation}
This means function \( \nabla F_m(\boldsymbol{w}_m^s, \boldsymbol{w}_m^u) \) is also $L_s$-Lipschitz continuous with respect to $\boldsymbol{w}_m^s$. The proofs for the other three properties are similar and are omitted here. Then we have
\begin{equation}
\begin{aligned}
&F_{m}(\boldsymbol{w}_{m,t+1}^{s},\boldsymbol{w}_{m,t+1}^{u})-F_{m}(\boldsymbol{w}_{m,t}^{s},\boldsymbol{w}_{m,t+1}^{u})\\
&\overset{(a)}\leq\left\langle\nabla_{s}F_{m}(\boldsymbol{w}_{m,t}^{s},\boldsymbol{w}_{m,t+1}^{u}),\boldsymbol{w}_{m,t+1}^{s}-\boldsymbol{w}_{m,t}^{s}\right\rangle\\
&\hspace{0.5cm}+\frac{L_{s}}{2}\left\|\boldsymbol{w}_{m,t+1}^{s}-\boldsymbol{w}_{m,t}^{s}\right\|^{2}\\
&\overset{(b)}\leq\left\langle\nabla_{s}F_{m}(\boldsymbol{w}_{m,t}^{s},\boldsymbol{w}_{m,t+1}^{u}),\boldsymbol{w}_{m,t+1}^{s}-\tilde{\boldsymbol{w}}_{m,t+1}^{s}\right\rangle\\
&\hspace{0.5cm}+\left\langle\nabla_{s}F_{m}(\boldsymbol{w}_{m,t}^{s},\boldsymbol{w}_{m,t+1}^{u}),\tilde{\boldsymbol{w}}_{m,t+1}^{s}-\boldsymbol{w}_{m,t}^{s}\right\rangle\\
&\hspace{0.5cm} +\frac{L_{s}}{2}\left\|\boldsymbol{w}_{m,t+1}^{s}-\boldsymbol{w}_{m,t}^{s}\right\|^{2}
\end{aligned}
\end{equation}
where (a) follows from the \( L_s \)-smooth property of \( F_m(\boldsymbol{w}_m^s, \boldsymbol{w}_m^u) \), and (b) results from adding and subtracting the term \( \tilde{\boldsymbol{w}}_{m,t+1}^{s} \). Also, since \( F_m(\boldsymbol{w}_m^s, \boldsymbol{w}_m^u) \) is $L_u$-smooth with respect to $\boldsymbol{w}_m^u)$, we have
\begin{equation}
\begin{aligned}
&F_{m}(\boldsymbol{w}_{m,t}^{s},\boldsymbol{w}_{m,t+1}^{u})-F_{m}(\boldsymbol{w}_{m,t}^{s},\boldsymbol{w}_{m,t}^{u})\\
&\leq\left\langle\nabla_{u}F_{m}(\boldsymbol{w}_{m,t}^{s},\boldsymbol{w}_{m,t}^{u}),\boldsymbol{w}_{m,t+1}^{u}-\boldsymbol{w}_{m,t}^{s}\right\rangle\\
&\hspace{4cm} +\frac{L_{u}}{2}\left\|\boldsymbol{w}_{m,t+1}^{u}-\boldsymbol{w}_{m,t}^{u}\right\|^{2}
\end{aligned} 
\end{equation}
Adding (59) and (60) together, we get
\begin{equation}
\begin{aligned}
&F_{m}(\boldsymbol{w}_{m,t+1}^{s},\boldsymbol{w}_{m,t+1}^{u})-F_{m}(\boldsymbol{w}_{m,t}^{s},\boldsymbol{w}_{m,t}^{u})\\
&\leq\left\langle\nabla_{s}F_{m}(\boldsymbol{w}_{m,t}^{s},\boldsymbol{w}_{m,t+1}^{u}),\boldsymbol{w}_{m,t+1}^{s}-\tilde{\boldsymbol{w}}_{m,t+1}^{s}\right\rangle\\
&\hspace{0.5cm} +\left\langle\nabla_{s}F_{m}(\boldsymbol{w}_{m,t}^{s},\boldsymbol{w}_{m,t+1}^{u}),\tilde{\boldsymbol{w}}_{m,t+1}^{s}-\boldsymbol{w}_{m,t}^{s}\right\rangle\\
&\hspace{0.5cm}+\left\langle\nabla_{u}F_{m}(\boldsymbol{w}_{m,t}^{s},\boldsymbol{w}_{m,t}^{u}),\boldsymbol{w}_{m,t+1}^{u}-\boldsymbol{w}_{m,t}^{u}\right\rangle\\
&\hspace{0.5cm}+\frac{L_{s}}{2}\left\|\boldsymbol{w}_{m,t+1}^{s}-\boldsymbol{w}_{m,t}^{s}\right\|^{2}+\frac{L_{u}}{2}\left\|\boldsymbol{w}_{m,t+1}^{u}-\boldsymbol{w}_{m,t}^{u}\right\|^{2}
\end{aligned}
\end{equation}
This completes the proof.
\subsection{Proof of Theorem 1}\label{AppendixB}
To facilitate the derivation, we introduce two auxiliary variables $\hat{\boldsymbol{w}}_{m,t+1}^{s}=\boldsymbol{w}_{m,t}^{s}-\eta\frac{\sum_{n= 1}^ND_n\boldsymbol{G}_{m,n,t}^s}D $ and  
$\tilde{\boldsymbol{w}}_{m,t+1}^{s}=\boldsymbol{w}_{m,t}^{s}-\frac{\sum_{n=1}^N\beta_{m,n}^tD_n\boldsymbol{G}_{m,n,t}}{\sum_{n=1}^N\beta_{m,n}^tD_n}$.
For the term $\left\langle\nabla_{s}F_{m}(\boldsymbol{w}_{m,t}^{s},\boldsymbol{w}_{m,t+1}^{u}),\tilde{\boldsymbol{w}}_{m,t+1}^{s}-\boldsymbol{w}_{m,t}^{s}\right\rangle$, it can be derived that
\begin{align}
&\left\langle\nabla_{s}F_{m}(\boldsymbol{w}_{m,t}^{s},\boldsymbol{w}_{m,t+1}^{u}),\tilde{\boldsymbol{w}}_{m,t+1}^{s}-\boldsymbol{w}_{m,t}^{s}\right\rangle\notag\\
&=\left\langle\nabla_{s}F_{m}(\boldsymbol{w}_{m,t}^{s},\boldsymbol{w}_{m,t}^{u}),\tilde{\boldsymbol{w}}_{m,t+1}^{s}-\boldsymbol{w}_{m,t}^{s}\right\rangle\notag\\
&\hspace{0.5cm} +\left\langle\nabla_{s}F_{m}(\boldsymbol{w}_{m,t}^{s},\boldsymbol{w}_{m,t+1}^{u})-\nabla_{s}F_{m}(\boldsymbol{w}_{m,t}^{s},\boldsymbol{w}_{m,t}^{u}),\right .\notag\\
&\hspace{6cm} \left .\tilde{\boldsymbol{w}}_{m,t+1}^{s}-\boldsymbol{w}_{t}^{s}\right\rangle\notag\\
&\leq\left\langle\nabla_{s}F_{m}(\boldsymbol{w}_{m,t}^{s},\boldsymbol{w}_{m,t}^{u}),\tilde{\boldsymbol{w}}_{m,t+1}^{s}-\boldsymbol{w}_{m,t}^{s}\right\rangle\notag\\
&\hspace{0.5cm}+ \left \| \nabla_{s}F_{m}(\boldsymbol{w}_{m,t}^{s},\boldsymbol{w}_{m,t+1}^{u})-\nabla_{s}F_{m}(\boldsymbol{w}_{m,t}^{s},\boldsymbol{w}_{m,t}^{u}) \right \| \notag\\
&\hspace{5.5cm} \times \left \|  \tilde{\boldsymbol{w}}_{m,t+1}^{s}-\boldsymbol{w}_{t}^{s}\right \|\notag\\
&\leq\left\langle\nabla_{s}F_{m}(\boldsymbol{w}_{t}^{s},\boldsymbol{w}_{m,t}^{u}),\tilde{\boldsymbol{w}}_{m,t+1}^{s}-\boldsymbol{w}_{t}^{s}\right\rangle\notag\\
&\hspace{0.5cm}+ \sqrt{L_{su}} \left \| \boldsymbol{w}_{m,t+1}^{u}-\boldsymbol{w}_{m,t}^{u}\right \|\sqrt[]{L_{su}}\left \|  \tilde{\boldsymbol{w}}_{m,t+1}^{s}-\boldsymbol{w}_{t}^{s}  \right \|\notag\\
&\leq \left\langle\nabla_{s}F_{m}(\boldsymbol{w}_{t}^{s},\boldsymbol{w}_{m,t}^{u}),\tilde{\boldsymbol{w}}_{m,t+1}^{s}-\boldsymbol{w}_{t}^{s}\right\rangle\notag\\
&\hspace{0.5cm}+\frac{1}{2}L_{su}\left \| \boldsymbol{w}_{m,t+1}^{u}-\boldsymbol{w}_{m,t}^{u}\right \|^2+\frac{1}{2}L_{su}\left \|  \tilde{\boldsymbol{w}}_{m,t+1}^{s}-\boldsymbol{w}_{t}^{s}  \right \|^2
\end{align}
For ease of representation, we let $L_1=2L_{s}+L_{su}$, $L_2=L_{u}+L_{su}$ and $\boldsymbol{w}_{m,t}=(\boldsymbol{w}_{m,t}^{s},\boldsymbol{w}_{m,t}^{u})$ in the following proof. Now, we have
\begin{equation}
\begin{aligned}
&F_{m}(\boldsymbol{w}_{t+1}^{s},\boldsymbol{w}_{m,t+1}^{u})-F_{m}(\boldsymbol{w}_{t}^{s},\boldsymbol{w}_{m,t}^{u})\\
&\leq\underbrace{\left\langle\nabla_{s}F_{m}(\boldsymbol{w}_{m,t}^{s},\boldsymbol{w}_{m,t+1}^{u}),\boldsymbol{w}_{m,t+1}^{s}-\tilde{\boldsymbol{w}}_{m,t+1}^{s}\right\rangle}_A\\
&\hspace{4cm}+\underbrace{L_{su}\left\|\boldsymbol{w}_{m,t+1}^{s}-\tilde{\boldsymbol{w}}_{m,t+1}^{s}\right\|^{2}}_B\\
&+\underbrace{\left\langle\nabla_{s}F_{m}(\boldsymbol{w}_{m,t}), \tilde{\boldsymbol{w}}_{m,t+1}^{s} \!-\! \boldsymbol{w}_{m,t}^{s}\right\rangle}_C
+\underbrace{\frac{L_1}{2}\left\|\tilde{\boldsymbol{w}}_{m,t+1}^{s}-\boldsymbol{w}_{m,t}^{s}\right\|^{2}}_D\\
&+\underbrace{\left\langle \nabla_{u} F_{m}(\boldsymbol{w}_{m,t}), \boldsymbol{w}_{m,t+1}^{u} \!-\! \boldsymbol{w}_{m,t}^{u} \right\rangle}_E
+\underbrace{\frac{L_2}{2}\left\|\boldsymbol{w}_{m,t+1}^{u}-\boldsymbol{w}_{m,t}^{u}\right\|^{2}}_F\\
\end{aligned}
\end{equation}
In the following, we will bound these six items separately.
For $A+B$, it is easy to get
\begin{equation}
\begin{aligned}
&A+B\\
&\leq (\frac{1}{2} +L_{su})\left\|\boldsymbol{w}_{m,t+1}^{s}-\tilde{\boldsymbol{w}}_{m,t+1}^{s}\right\|^{2}+\epsilon _s^2\\
&\leq (1+2L_{su})\underbrace{\left\|\boldsymbol{w}_{m,t+1}^{s}-\hat{\boldsymbol{w}}_{m,t+1}^{s}\right\|^{2}}_{A_1}\\
&\hspace{0.5cm}+(1+2L_{su})\underbrace{\left\|\hat{\boldsymbol{w}}_{m,t+1}^{s}-\tilde{\boldsymbol{w}}_{m,t+1}^{s}\right\|^{2}}_{A_2}+\epsilon_s^2
\end{aligned}
\end{equation}

Now, we bound $A_1$ and $A_2$. For the term $A_1=\left \|\boldsymbol{w}_{m,t+1}^{s}-\hat {\boldsymbol{w}}_{m,t+1}^{s}\right \|^2$, we can derive that

\begin{align}
&\left \|\boldsymbol{w}_{m,t+1}^{s}-\hat {\boldsymbol{w}}_{m,t+1}^{s}  \right \|^2\notag\\
&=\left \|\sum_{i\in S_{m,t}}\sum_{n = 1}^N\frac{\beta_{i,n}^{t}D_i\boldsymbol{G}_{i,n,t}^s}{\sum_{i
\in S_{m,t}}D_i}-\frac{\sum_{n= 1}^ND_n\boldsymbol{G}_{m,n,t}^s}D\right \|^2\notag\\
&=\left \|\frac{\sum_{i\in S_{m,t}}\sum_{n = 1}^N\beta_{i,n}^{t}D_i(\boldsymbol{G}_{i,n,t}^s-\boldsymbol{G}_{m,n,t}^s)}{\sum_{i\in S_{m,t}}D_i}\right.\notag\\
&\hspace{0.5cm} \left .+\frac{\sum_{i\in S_{m,t}}\sum_{n = 1}^N\beta_{i,n}^{t}D_i\boldsymbol{G}_{m,n,t}^s}{\sum_{i\in S_{m,t}}D_i}-\frac{\sum_{n= 1}^ND_n\boldsymbol{G}_{m,n,t}^s}D\right \|^2\notag\\
&\leq 2\left \|\frac{\sum_{i\in S_{m,t}\setminus m}\sum_{n = 1}^N\beta_{i,n}^{t}D_i(\boldsymbol{G}_{i,n,t}^s-\boldsymbol{G}_{m,n,t}^s)}{\sum_{i\in S_{m,t}}D_i}\right\|^2\notag\\
&\hspace{0.5cm} +2\left \|(\frac{1}{\sum_{i\in S_{m,t}}D_i}-\frac{1}{D}  )\sum_{i\in S_{m,t}}\sum\nolimits_{n = 1}^N\beta_{i,n}^{t}D_i\boldsymbol{G}_{m,n,t}^s\right .\notag\\
&\hspace{0.5cm}+ \left. \frac{\sum_{i\in \mathcal{N}\setminus S_{m,t}}\sum_{n = 1}^N\beta_{i,n}^{t}D_i\boldsymbol{G}_{m,n,t}^s}{D} \right \|^2\notag\\
&\leq 2\left \|\frac{\sum_{i\in S_{m,t}\setminus m}\sum_{n = 1}^N\beta_{i,n}^{t}D_i(\boldsymbol{G}_{i,n,t}^s-\boldsymbol{G}_{m,n,t}^s)}{\sum_{i\in S_{m,t}}D_i}\right\|^2\notag\\
&\hspace{0.5cm}+8(1-\frac{\sum_{i\in S_{m,t}}D_{i,t}}{D} )^2\epsilon_s^2\notag\\
&\leq 4\left \|\frac{\sum_{i\in S_{m,t}\setminus m}\sum_{n = 1}^N\beta_{i,n}^{t}D_i(\boldsymbol{G}_{i,n,t}^s-\hat{\boldsymbol{G}}_{i,n,t}^s)}{\sum_{i\in S_{m,t}}D_i}\right\|^2\notag\\
&\hspace{0.5cm}+4\left \|\frac{\sum_{i\in S_{m,t}\setminus m}\sum_{n = 1}^N\beta_{i,n}^{t}D_i(\hat{\boldsymbol{G}}_{i,n,t}^s-\boldsymbol{G}_{m,n,t}^s)}{\sum_{i\in S_{m,t}}D_i}\right\|^2\notag\\
&\hspace{0.5cm}+8(1-\frac{\sum_{i\in S_{m,t}}D_{i,t}}{D} )\epsilon_s^2\notag\\
&\leq 4\left \|\frac{\sum_{i\in S_{m,t}\setminus m}\sum_{n = 1}^N\beta_{i,n}^{t}D_i(\hat{\boldsymbol{G}}_{i,n,t}^s-\boldsymbol{G}_{m,n,t}^s)}{\sum_{i\in S_{m,t}}D_i}\right\|^2\notag\\
&\hspace{0.5cm}+ 4(1-\frac{D_{m,t}}{\sum_{i\in S_{m,t}}D_{i,t}})^2\delta ^2+8(1-\frac{\sum_{i\in S_{m,t}}D_{i,t}}{D} )^2\epsilon_s^2\notag\\
&\leq 4(1\!-\!\frac{D_{m,t}}{\sum_{i\in S_{m,t}}D_{i,t}})^2(\delta^2\!+\!2\epsilon_s^2)+8(1\!-\!\frac{\sum_{i\in S_{m,t}}D_{i,t}}{D} )^2\epsilon_s^2
\end{align}

where $\hat{\boldsymbol{G}}_{i,n,t}^s$ means UAV $n$ perform the update for task $i$ by using the same initial parameter with task $m$.

For the term \( A_2=\left \|\hat{\boldsymbol{w}}_{m,t+1}^{s} - \tilde{\boldsymbol{w}}_{m,t+1}^{s} \right \|^2 \), by applying similar mathematical derivation as in case $A_1$, we obtain the following inequality:
\begin{equation}
\left \|\hat{\boldsymbol{w}}_{m,t+1}^{s} - \tilde{\boldsymbol{w}}_{m,t+1}^{s} \right \|^2 \leq 4\left( 1 - \frac{D_{m,t}}{D}  \right)^2\epsilon_s^2
\end{equation}
The details of the derivation are omitted here for simplicity. To facilitate the proof of $C+D$, we introduce auxiliary variables
\begin{equation}
\xi_{m,t}^s=\frac{\sum_{n=1}^N\beta_{m,n}^tD_n\boldsymbol{G}_{m,n,t}}{\sum_{n=1}^N\beta_{m,n}^tD_n}
\end{equation}
Then it can be derived that
\begin{align}
&\left\langle\nabla_{s}F_{m}(\boldsymbol{w}_{m,t}),\tilde{\boldsymbol{w}}_{m,t+1}^{s}-\boldsymbol{w}_{t}^{s}\right\rangle
+\frac{L_1}{2}\left\|\tilde{\boldsymbol{w}}_{m,t+1}^{s}-\boldsymbol{w}_{m,t}^{s}\right\|^{2}\notag\\
&=-\eta\left\langle\nabla_{s}F_{m}(\boldsymbol{w}_{m,t})), \xi_{m,t}^s-\nabla_{s}F_{m}(\boldsymbol{w}_{m,t})\right.\notag\\
&\hspace{3cm} \left.+\nabla_{s}F_{m}(\boldsymbol{w}_{m,t})\right\rangle+\frac{L_1\eta ^2}2\|\xi_{m,t}^s\|^2\notag\\
&\leq -\frac{\eta}{2}\|\nabla_sF_m(\boldsymbol{w}_{m,t})\|^2+\frac{\eta}2E\|\xi_{m,t}^s-\nabla_{s}F_{m}(\boldsymbol{w}_{m,t})\|^{2}\notag\\
&\hspace{0.5cm} +\frac{L_1\eta^2}{2} E\|\xi_{m,t}^s-\nabla_{s}F_{m}(\boldsymbol{w}_{m,t})+\nabla_{s}F_{m}(\boldsymbol{w}_{m,t})\|^{2}\notag\\
&\leq (-\frac{\eta}{2}+L_1\eta ^2)E\parallel\nabla_{s}F_{m}(\boldsymbol{w}_{m,t})\parallel^{2}\notag\\
&\hspace{0.5cm}+(\frac{\eta}{2}+L_1\eta^2 )\left\|\frac{\sum\nolimits_{i\in \mathcal{M}\setminus m}\sum_{i=1}^N\beta_{i,n}^{t}D_nG_{i,n,t}}{D}\right.\notag\\
&\hspace{0.5cm}\left. +\frac{\sum\nolimits_{i=1}^N\beta_{m,n}^tD_nG_{m,n,t}}{D}-\frac{\sum\nolimits_{n=1}^{N}\beta_{m,n}^{t}D_{n}G_{m,n,t}}{\sum\nolimits_{n=1}^{N}\beta_{m,n}^{t}D_{n}}\right \|^2\notag\\
&\leq (-\frac{\eta}{2}+L_1\eta ^2)E\parallel\nabla_{s}F_{m}(\boldsymbol{w}_{m,t})\parallel^{2}\notag\\
&\hspace{0.5cm}+(\frac{\eta}{2}+L_1\eta^2 )(1-\frac{\sum_{i=1}^N\beta_{m,n}^tD_n}{D} )^2\epsilon_s^2
\end{align}
Therefore, the above inequality holds when \(\eta < \frac{1}{2(2L_{s} + L_{su})}\).
Similarly, for the term $E+F$ we can prove that when $\eta < \frac{1}{2(L_{u}+L_{su})}$, the following inequality holds
\begin{equation}
\begin{aligned}
&\left\langle\nabla_{u}F_{m}(\boldsymbol{w}_{m,t}),\tilde{\boldsymbol{w}}_{m,t+1}^{u}-\boldsymbol{w}_{t}^{u}\right\rangle
+\frac{L_2}{2}\left\|\tilde{\boldsymbol{w}}_{m,t+1}^{u}-\boldsymbol{w}_{m,t}^{u}\right\|^{2}\\
&\leq (-\frac{\eta}{2}+L_2\eta ^2)E\parallel\nabla_{u}F_{m}(\boldsymbol{w}_{m,t})\parallel^{2}\\
&\hspace{0.5cm}+(\frac{\eta}{2}+L_2\eta^2 )(1-\frac{\sum_{i=1}^N\beta_{m,n}^tD_n}{D} )^2\epsilon_u^2
\end{aligned}
\end{equation}
The details are omitted here for simplicity. Finally, substituting the terms $A-F$ into (63), we get that when $\eta < \min\left(\frac{1}{2(2L_{s} + L_{su})},\frac{1}{2(L_{u} + L_{su})}\right)$,
\begin{equation}
\begin{aligned}
&F_{m}(\boldsymbol{w}_{m,t+1})-F_{m}(\boldsymbol{w}_{m,t})\\
&\leq (-\frac{\eta }{2}+L\eta ^2) \left [  \mathbb{E}\left \| \nabla_{s}F_{m}(\boldsymbol{w}_{m,t})\right \|^2+ \mathbb{E}\left \| \nabla_{u}F_{m}(\boldsymbol{w}_{m,t})\right \|^2\right ]\\
&\hspace{0.5cm} +4(1+2L_{su})(1-\frac{D_{m,t}}{\sum_{i\in S_{m,t}}D_{i,t}})^2\delta^2\\
&\hspace{0.5cm}+8(1+2L_{su})(1\!-\!\frac{\sum_{i\in S_{m,t}}D_{i,t}}{D} )^2\epsilon_s^2+\Omega_t
\end{aligned}
\end{equation}

where $ \Omega_t=(5+8L_{su}+\eta(2\eta L_s+\eta L_{su}+\frac{1}{2} ))(1-\frac{D_{m,t}}{D} )^2\epsilon _s^2+(\frac{\eta}{2}+\eta^2(L_u+L_{su}) )(1-\frac{D_{m,t}}{D} )^2\epsilon_u^2$.

\subsection{Proof of Theorem 2}\label{AppendixC}
According to lemma \hyperref[lemma2]{2}, we have
\begin{equation}
\Delta\left(\boldsymbol{Q}^{t}\right)\leq B  + \sum\nolimits_{n=1}^N Q_{n,t} \left(E_{n,t}  - \bar{E}_n \right)
\end{equation}
It follows that
\begin{equation}\label{eq71}
\Delta(Q^{t})-VU^{t}\leq B+\sum_{n=1}^{N}Q_{n,t}(E_{n,t}-\bar{E}_n)-VU^{t}
\end{equation}
\hyperref[eq71]{(71)} imposes no requirements on the strategy. Specifically, under the conditions $\alpha_{m}^t = 0,\forall m$ and $E_{n,t} = 0,\forall n$, we have:
\begin{equation}
\Delta(Q^{t})-V\hat{U}^{t}\leq B-\sum_{n=1}^{N}Q_{n,t}\bar{E}_n\leq B
\end{equation}
Let $U^{t,*}$ represent the maximum utility achievable by the optimal offline policy in round $t$ under no energy constraints. From the above inequality, we derive:
\begin{equation}
L[Q^{t+1}]-L[Q^t]\leq B+VU^{t,*}
\end{equation}
Summing it up from $t = 0$ to $t = T - 1$, we obtain:
\begin{equation}
L[Q^{T}]-L[Q^0]\leq TB+V \sum_{t=0}^{T}  U^{t,*}
\end{equation}
\begin{equation}
\sum_{n=1}^{N}Q_{n,T}^{2}\leq2TB+2V\sum_{t=0}^{T-1} U^{t,*}
\end{equation}
Then we have
\begin{equation}
\sum_{n=1}^{N}Q_{n,T}\leq \sqrt{N\sum_{n=1}^{n}Q_{n,T}^2}\leq\sqrt{2NTB+2NV\sum_{t=0}^{T-1} U^{t,*}}
\end{equation}
From \hyperref[eq20]{(20)},we know $Q_{n,t+1} > Q_{n,t} + E_{n,t} - \bar{E}_n$, that is $E_{n,t} - \bar{E}_n \leq Q_{n,t+1} - Q_{n,t}$, summing up from $t=0$ to $t=T-1$ and from $n=1$ to $n=N$, we have
\begin{equation}
\sum_{t=0}^{T-1}\sum_{n=1}^{N} \left(E_{n,t} - \bar{E}_n\right ) \leq \sum_{n=1}^{N}Q_{n,T} \leq \sqrt{2TB+2V \sum_{t=0}^{T-1}U_{max}^t} 
\end{equation}
Rearrange the above formula, we get
\begin{equation}
\frac{1}{T}\sum_{t=0}^{T-1}\sum_{n=1}^{N}E_{n,t} \leq \sum_{n=1}^{N}\bar{E}_n+   \sqrt{\frac{2B}{T}+\frac{2V\sum_{t=0}^{T-1}U^{t,*}}{T^2}} 
\end{equation}
From \hyperref[eq25]{(25)}, we have
\begin{equation}
\Delta(Q^t) - V U^t \leq B + \sum_{n=0}^NQ_{n,t} (E_{n,t} - \bar{E}_n) - V U^t
\end{equation}
Denoting $\hat{U}^t$ as the achieved utility by our algorithm in round $t$, sine problem $\mathcal{P}_2$ is minimized in each round, there is no other policy can get a lower objective function, so we have
\begin{equation}
\Delta(Q^t) - V \hat{U}^t \leq B + \sum_{n=0}^NQ_{n,t} (E_{n,t}^* - \bar{E}_n) - V U^{t,*}
\end{equation}
Summing it up from $t=0$ to $t=T-1$, then
\begin{equation}
-\sum_{t=0}^{T} V\hat{U}^t \leq TB +\sum_{n=0}^{N} \sum_{t=0}^{T-1} Q_{n,t} (E_{n,t}^* - \bar{E}_n) - \sum_{t=0}^{T-1} VU^{t,*}
\end{equation}
Next, we bound
\begin{equation}
\begin{aligned}
\sum_{n=0}^{N} \sum_{t=0}^{T-1} Q_{n,t} (E_{n,t}^* - \bar{E}_n)&= \sum_{t=0}^{T-1} \sum_{n=1}^{N} (Q_{n,t} - Q_{n,0}) (E_{n,t}^* - \bar{E}_n)\\
\leq 2&B\sum_{t=0}^{T-1}t = TB (T-1)
\end{aligned}
\end{equation}
Then we have
\begin{equation}
-\sum_{t=0}^{T} V\hat{U}^t \leq TB + TB(T-1) - \sum_{t=0}^{T} VU^{t,*}
\end{equation}
Rearrange the formula, the proof is completed.

\bibliographystyle{IEEEtran}  
\bibliography{reference}  
\end{document}